\documentclass[10pt,twocolumn,letterpaper]{article}

\usepackage[final]{cvpr}              % To produce the CAMERA-READY version
\usepackage{graphicx}	
\usepackage{amsmath}	
\usepackage{amssymb}	
\usepackage{booktabs}
\usepackage{times}
\usepackage{microtype}
\usepackage{epsfig}
\usepackage{caption}
\usepackage{float}
\usepackage{placeins}
\usepackage{color, colortbl}
\usepackage{stfloats}
\usepackage{enumitem}
\usepackage{tabularx}
\usepackage{xstring}
\usepackage{multirow}
\usepackage{xspace}
\usepackage{url}
\usepackage{subcaption}
\usepackage{xcolor}
\usepackage[hang,flushmargin]{footmisc}
\usepackage{hyperref}

% ...

\newcommand{\R}[1]{{%
    \textbf{%
        \ifstrequal{#1}{1}{\textcolor{red}{R#1}}{%
        \ifstrequal{#1}{2}{\textcolor{blue}{R#1}}{%
        \ifstrequal{#1}{3}{\textcolor{magenta}{R#1}}{%
        \ifstrequal{#1}{4}{\textcolor{teal}{R#1}}{%
                           \textcolor{cyan}{R#1}%
        }}}}%
    }%
}}

\usepackage{xspace}
\makeatletter
\DeclareRobustCommand\onedot{\futurelet\@let@token\@onedot}
\def\@onedot{\ifx\@let@token.\else.\null\fi\xspace}
\def\eg{\emph{e.g}\onedot} \def\Eg{\emph{E.g}\onedot}
\def\ie{\emph{i.e}\onedot} \def\Ie{\emph{I.e}\onedot}
\def\cf{\emph{c.f}\onedot} 
\def\etc{\emph{etc}\onedot}

\usepackage{pifont}%
\newcommand{\cmark}{\textcolor{teal}{\ding{51}}}%
\newcommand{\xmark}{\textcolor{red}{\ding{55}}}%
 \newcommand{\textapproxx}{{\raise.17ex\hbox{$\scriptstyle\mathtt{\sim}$}}}
 
\newcolumntype{P}[1]{>{\centering\arraybackslash}p{#1}}

\newcommand{\ours}{\texttt{ALC-ITR}}
\newcommand{\modelname}{RRA-VL}
\newcommand{\problemname}{LC-MMR}

\title{Anatomy-Aware Conditional Image-Text Retrieval}

\author{Meng Zheng$^{1}$, Jiajin Zhang$^{2}$\thanks{This work was done during the internship of Jiajin Zhang at United Imaging Intelligence, Boston, MA.}, Benjamin Planche$^{1}$, Zhongpai Gao$^{1}$, Terrence Chen$^{1}$, Ziyan Wu$^{1}$
\\$^{1}$United Imaging Intelligence, Boston, MA~$^{2}$ Rensselaer Polytechnic Institute, Troy, NY
\\{$^{1}$\{firstname.lastname\}@uii-ai.com}, $^{2}$zhangj41@rpi.edu
}
\date{}

\begin{document}
\maketitle
\begin{abstract}
% Image-Text Retrieval (ITR) finds broad applications in healthcare, aiding clinicians and radiologists by automatically retrieving relevant patient cases in the database given the query image and/or report, for more efficient clinical diagnosis and treatment planning. 
%Conventional class-based (disease-level) medical Image-Text Retrieval (ITR) systems typically only rely on global image or text representations for measuring patient image/report similarities, which overlook local distinctiveness across patient cases.

Image-Text Retrieval (ITR) finds broad applications in healthcare, aiding clinicians and radiologists by automatically retrieving relevant patient cases in the database given the query image and/or report, for more efficient clinical diagnosis and treatment, especially for rare diseases. However conventional ITR systems typically only rely on global image or text representations for measuring patient image/report similarities, which overlook local distinctiveness across patient cases.
% considering the fact that: 1) concurrent diseases may happen at same or different anatomical regions for the same patient, 2) pathologies typically only happen at small regions of a medical image. Learning global image or textual representations thus ignores anatomical region-level information 
This often results in suboptimal retrieval performance. In this paper, we propose an Anatomical Location-Conditioned Image-Text Retrieval (\ours) framework, which, given a query image and the associated suspicious anatomical region(s), aims to retrieve similar patient cases exhibiting the same disease or symptoms in the same anatomical region. To perform location-conditioned multimodal retrieval,
% \bpnote{\ours\ is described as a "system", so a bit unclear what "to perform [the system]" means} \mznote{how about this? I was a bit worried too many abbreviations might cause confusion.}\bpnote{The sentence is better IMO. We could remove the acronym though -- I guess it doesn't need to be introduced in the Abstract yet?}
we learn a medical Relevance-Region-Aligned Vision Language (\modelname) model with semantic global-level and region-/word-level alignment to produce generalizable, well-aligned multi-modal representations. Additionally, we perform location-conditioned contrastive learning to further utilize cross-pair region-level contrastiveness for improved multi-modal retrieval. We show that our proposed \modelname~achieves state-of-the-art localization performance in phase-grounding tasks, and satisfying multi-modal retrieval performance with or without location conditioning. Finally, we thoroughly investigate the generalizability and explainability of our proposed \ours~system in providing explanations and preliminary diagnosis reports given retrieved patient cases (conditioned on anatomical regions), with proper off-the-shelf LLM prompts. 
\end{abstract}    
\section{Introduction}
\label{sec:intro}

Image-text Retrieval (ITR) has been a widely studied problem in the AI research community \cite{retrieval_IJCAI22,ALBEF_NIPS21,ROSITA_MM21,DIME_SIGIR21}. It aims to retrieve relevant images or text from a large database, given a query image or text description. Retrieval tasks have broad applications, particularly in healthcare \cite{IR_NC17,LargescaleRF_MIA18,XTRA_IPMI23,XAIretrieval_WACV22,pmlr_MLH21,CBIR_MICCAI21}, supporting clinical diagnosis, treatment planning and disease monitoring, by automatically retrieving and referencing relevant cases from past patient databases. 
Recently, foundation models like Visual Language Models (VLMs) and Large Language Models (LLMs), have been widely explored in multi-modality (image-text) or cross-modality (image$\rightarrow$text or text$\rightarrow$image) retrieval tasks \cite{ViSTAVA_CVPR22,MSCXR_ECCV22,XTRA_IPMI23,Pan_2023_CVPR} due to their superior zero-shot transferability in performing downstream tasks like classification and retrieval, achieved by learning generalizable semantic-aligned textual and visual representations.

\begin{figure*}[h!]
    \centering
    \includegraphics[width=0.95\linewidth]{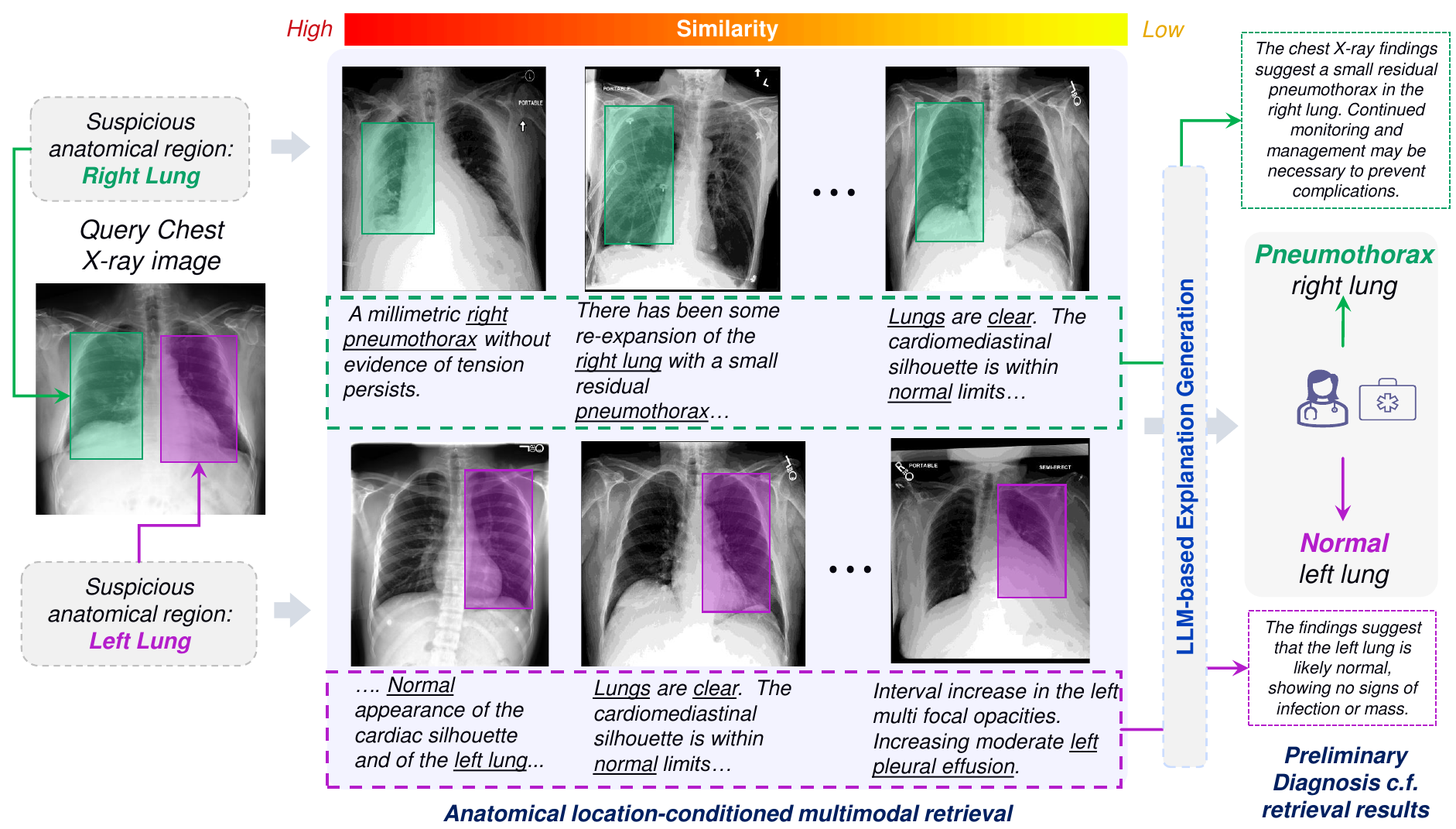}
    \caption{Proposed multimodal retrieval system (\ours) with anatomical region conditioning. Given a query chest X-ray image and a suspicious anatomical region, our proposed system retrieves the most relevant patient cases showing the same or similar disease, for the same anatomical regions. The system then prompts LLMs based on retrieval results for explanation generation and preliminary diagnosis. }
    \vspace{-0.4cm}
    \label{fig:teaser}
\end{figure*}

% Problem of Conditioned Image/Text Retrieval task
Though existing VLM-based solutions have demonstrated satisfactory performance in class-based retrieval tasks in medical domain \cite{GLoRIA_ICCV21,MSCXR_ECCV22,wei2023masked_arxiv}---\ie, retrieving similar patient cases with the same or similar type(s) of disease given a query medical image or associated report---the performed multi-/cross-modality retrieval primarily relies on global visual or textual report information. This may overlook local anatomical-region-level distinctiveness across patient cases leading to suboptimal retrieval performance, as certain diseases manifest differently depending on their anatomical locations. 
For instance, a mass in the lung apex might suggest Pancoast tumor, while one near the hilum might indicate metastasis or lymphadenopathy. 
On the other hand, as lesions typically occupy only small fractions of the image, a global-level retrieval may fail to account for these small region-specific differences, compromising retrieval accuracy. 

To overcome these limitations, we propose a novel system that retrieves relevant patient cases showing similar lesion
% \mznote{or lesion?}
\textit{conditioned on anatomical regions}, providing more effective and efficient diagnosis support to clinicians or radiologists. 
% \mznote{is the example convincing enough? Do we want to add argument saying that multiple diseases can happen at different regions?}\bpnote{if you mean the Pancoast/lymphadenopathy example, I think it is good. Mentioning the opposite cases (same disease in different regions) could be good if we have the space, but maybe not critical (?)/}
More specifically, as shown in Figure~\ref{fig:teaser}, we explore the problem of anatomical Location-Conditioned Multimodal Retrieval (\problemname). Given a query medical image and its region of interest—typically a suspicious anatomical area where radiologists or clinicians have significant uncertainty regarding the presence of a disease—our objective is to retrieve relevant patient cases from the database that exhibit the same or similar conditions \textit{at the same anatomical location}. 
% \mznote{shall I add: This is particularly helpful for rare disease...?}
Furthermore, leveraging the retrieved patient metadata, we prompt general LLMs (without domain-specific fine-tuning) to generate explanations and provide a preliminary diagnosis for the query case at the anatomical region level.

To develop an effective system for location-conditioned retrieval (\problemname), 
% as shown in Figure~\ref{fig:teaser},
directly applying existing medical VLM solutions~\cite{GLoRIA_ICCV21,MSCXR_ECCV22,wei2023masked_arxiv} trained on image-report pairs is suboptimal. These models optimize a shared embedding space mapped from \textit{global} visual and textual features, lacking explicit location conditioning. As a result, they may fail to capture critical regional relevance information, which is essential for the proposed \problemname~task.
% Learning local-aligned image-text representation for medical data is extremely challenging as most of the large-scale multi-modality datasets, \eg MIMIC-III~\cite{MIMIC_SD16} and PubMed~\cite{pubmed_MG12}, do not contain location-relevant annotations like bounding boxes and segmentation masks, as well as the local visual and textual correspondence information. 
While numerous works have been proposed \cite{muellerChEx_ECCV24,LIMITR_2023_ICCV,chen_finegrained_ACL24,GLoRIA_ICCV21,mutual_MICCAI21,pathology_MICCAI22} to improve the granularity of visual-textual alignment for better multi-modal representation learning, they are essentially learning global representations with vague region-level information. They are designed for specific downstream applications such as image classification, report generation, or pathology recognition; and thus do not properly transfer to \problemname.
% location-conditioned multi-modal retrieval.

To this end, we propose to learn a weakly-supervised region-relevance-aligned biomedical Vision Language (\modelname)~model 
% \mznote{change a name?} 
for \problemname. Specifically we focus on chest X-Ray (CXR) images paired with radiology reports; leveraging MIMIC-III/CXR~\cite{MIMIC_SD16}, a large-scale, multi-modal dataset with disease-level annotations which is well-suitable for our study. 
In addition to semantic global alignment, we extend \modelname with a novel region-/word-level visual-textual alignment module performing residual-connected intra- and cross-modality attention to enforce local alignment between medical images and reports. 
Furthermore, we enforce a location-conditioned contrastive learning loss, utilizing patient case distinctiveness at lesion level for improved multi-modal retrieval. We show that our proposed \modelname~enables \problemname~and can achieve greater retrieval performance with location-informed contrastive learning. 
We further demonstrate the lesion-level explainability of proposed \ours~system in providing diagnosis suggestions given retrieved patient case gallery, which facilitates more comprehensive patient case understanding and interpretation. 
% \mznote{I added \problemname~abbreviations and replaced some of the previous words, is it confusing?}\bpnote{I personally find it a bit hard to remember what all the new acronyms mean, but that's just my own opinion.}

To summarize, our contributions are listed as follows:
\begin{itemize}
    \item We introduce a novel Anatomical Location-Conditioned Image-Text Retrieval system (\ours), which is able to retrieve similar patient cases with same disease or symptoms \textit{conditioned on anatomical locations}, enabling patient case understanding and diagnosis support at finer level compared to conventional medical retrieval system. 
    \item We propose a weakly-supervised region-relevance-aligned medical Vision Language (\modelname)~model, with attentive region-/word-level and global semantic alignment for generalizable multi-modal representation learning. We further enforce a location-conditioned triplet mining loss for \modelname~learning, resulting in improved multi-modal retrieval performance.
    \item Our proposed \ours~system enables region-level explainability and provides preliminary diagnosis at anatomical locations, without requiring specific text generator learned with domain-specific knowledge and annotations.
    \item We demonstrate the superior multi-modal representation learning capability of the proposed \modelname~on various downstream tasks, including phase-grounding, multi-modality retrieval with/without location conditioning. The proposed \modelname~achieves state-of-the-art phase-grounding and satisfying multi-modal retrieval performance compared to existing solutions.
\end{itemize}

\section{Related Work}
\label{sec:related}

\subsection{Medical Vision-Language Model}
With the ongoing development of Foundation Models, much effort has been put in learning generalizable Vision Language Models (VLMs) from large-scale paired image-text data \cite{contrastive_ECCV20,transfer_ICML21,unicoder_IJCNLP19,Li2019VisualBERTAS,ViLBERT_NIPS19,VLBERT_ICLR20}. 
Common approaches typically enforce contrastive objectives on CLIP-like architectures \cite{clip_ICML21} to learn a joint visual-textual embedding space. 
Training medical VLMs remains, however, challenging; as public medical image-report paired datasets are in much smaller scale than in-the-wild image-text datasets such as LAION-5B~\cite{LAION-5B}, COYO-700M~\cite{kakaobrain2022coyo-700m} \etc. 
% \bpnote{ImageNet is actually considered small nowadays -- maybe list a larger dataset used to train VLMs?}. 
% Multi-modal medical image-report datasets are even more rare, due to demanding annotation efforts and privacy concerns. 
Numerous works have been proposed \cite{MSCXR_ECCV22,GLoRIA_ICCV21,mutual_MICCAI21,jointlearn_ECCV22,granularity_NIPS22,muellerChEx_ECCV24,LIMITR_2023_ICCV,chen_finegrained_ACL24,pathology_MICCAI22} to learn more generalizable visual-textual representation for medical data, by enforcing different levels of vision-language alignment during large-scale pretraining. While some are aiming at learning generic multi-modal representations \cite{GLoRIA_ICCV21,mutual_MICCAI21,pathology_MICCAI22,jointlearn_ECCV22,granularity_NIPS22,LIMITR_2023_ICCV}, others are specifically curated for different downstream tasks, \eg, report generation \cite{interactive_CVPR23,chen_finegrained_ACL24}, which may not easily translate to the proposed location-conditioned image/text retrieval (\problemname) problem. 
In this paper, we propose a novel attentive word-region-level local alignment scheme enforced with semantically weighted cross-modality consistency. We show that the proposed alignment scheme achieves superior localization capability in phase grounding tasks compared to existing solutions, while enabling \problemname~with explicit region conditioning.

\subsection{Image-Text Retrieval}
Image-text retrieval is a rapidly advancing AI research area that focuses on developing methods to retrieve relevant information across modalities, such as finding an image based on a textual query or vice versa. 
% , with applications ranging from search engines and recommendation systems to more specialized tasks such as medical image retrieval and content-based multimedia indexing. 
We refer the readers to \cite{retrieval_IJCAI22} for a detailed survey on image-text retrieval. Conventional image or text retrieval refers to category/class-level retrieval, \ie, the retrieval of candidates in a database which share the same category as the query. Existing approaches~\cite{Zhang2020ContrastiveLO_ICLR20,GLoRIA_ICCV21,granularity_NIPS22} typically leverages global similarities between image and/or text features for retrieval. 
We argue that local distinctiveness across different patient cases is crucial and should be utilized for more accurate and specialized retrieval particularly in medical domain. 
Specifically, we extend existing multi-modal (image-text) retrieval systems and propose a novel anatomical location-conditioned image-text retrieval (\ours) system, which retrieves relevant patient cases conditioned on anatomical regions, given the query radiology image and suspicious anatomical location(s). 
This further facilitates finer-level interpretability of a medical retrieval system by querying an image at various anatomical regions, and providing explanations and preliminary diagnosis accordingly based on region-level visual similarities.

\vspace{-0.1cm}
\section{Method}
\label{sec:method}
\begin{figure*}[h!]
    \centering
    \includegraphics[width=0.95\linewidth]{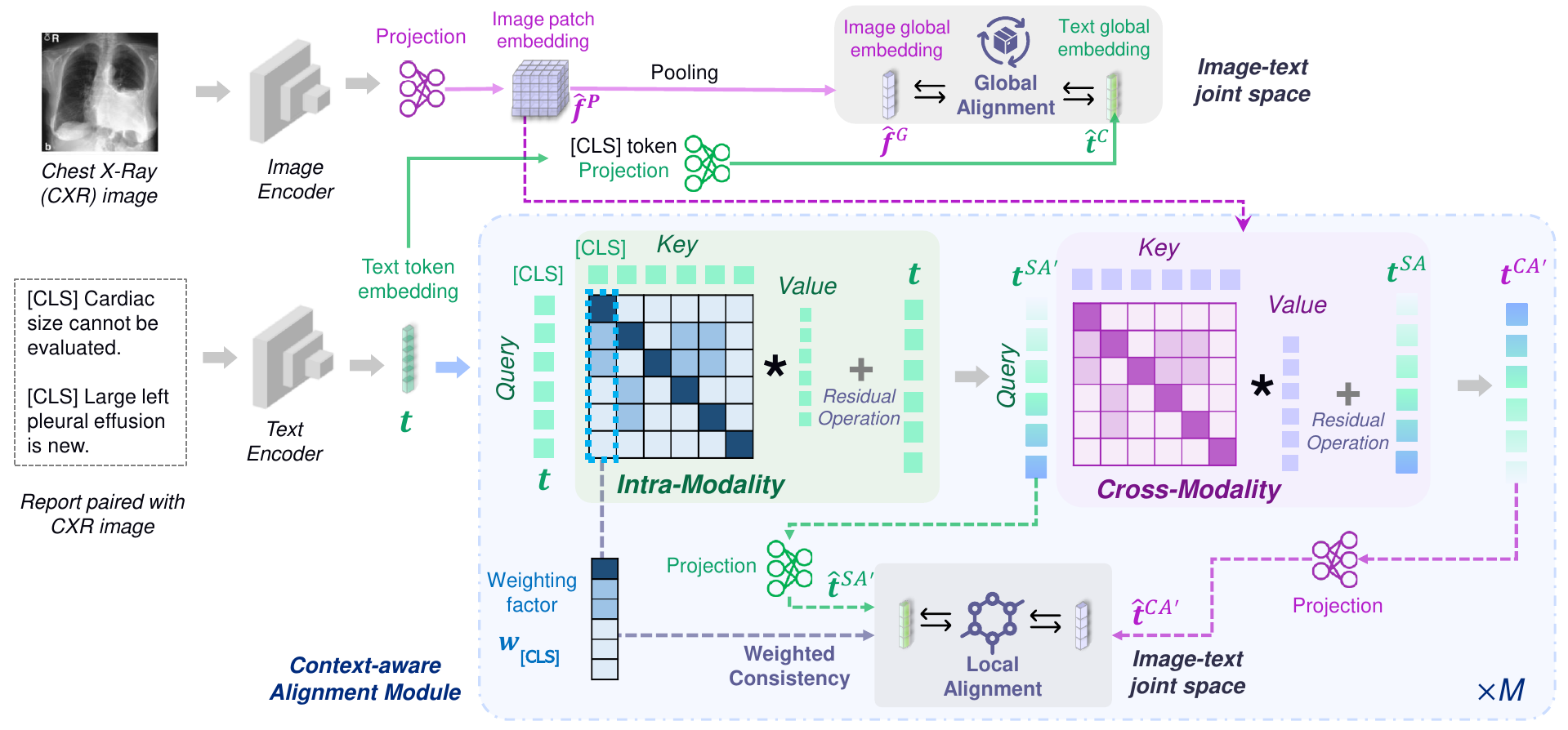}
    \caption{Proposed Region-Relevance-Aligned medical Vision Language (\modelname) model with global and local alignment.}
    \vspace{-0.4cm}
    \label{fig:pipeline}
\end{figure*}

To learn our proposed Region-Relevance-Aligned Vision Language (RRA-VL) model for Location-Conditioned Multimodal Retrieval (\problemname), we first achieve global and local semantic alignment of the image and text representations at multi-granularity levels in an unsupervised manner (Section~\ref{sec:method_alignment}). We then perform location-conditioned contrastive learning given disease labels as weak supervision for improved \problemname~(Section~\ref{sec:method_triplet}). We finally investigate the explainability of the proposed Anatomical Location-Conditioned Image-Text Retrieval (\ours) system in Section~\ref{sec:method_exp}. The overall pipeline is illustrated in Figure~\ref{fig:pipeline}.

\subsection{VLM Learning with Global/Local Alignment}
\label{sec:method_alignment}
\subsubsection{Feature Extraction}
\label{sec:align_pre}
Given a radiology image $I_i$, \eg a chest X-ray (CXR) image, along with its radiology report $T_i$, we adopt a ResNet-50~\cite{resnet_he2015} as the image encoder, $\mathbf{E}_I$, and a BERT model \cite{bert_NIPS17} as the text encoder $\mathbf{E}_T$, according to conventional medical vision-language model (VLM) frameworks \cite{MSCXR_ECCV22,GLoRIA_ICCV21,LIMITR_2023_ICCV}. Specifically, the image encoder takes in the CXR image $I_i$ and outputs a grid of local patch embeddings $\mathbf{{f}}^P_{i}$ of size $c\times w\times h$.
% the patch embeddings are then average-pooled to compute image global embedding $\mathbf{\hat{f}}^G_{i}$. 
The text encoder takes in the radiology report $T_i$ (paired with CXR image $I_i$) and generates a list of token representations $\mathbf{t}_{i}$. 
% with the first \texttt{[CLS]} token $\mathbf{\hat{t}}^{C}_{i}$ as the global textual representation. 
The local image patch embeddings and the textual token representations are then utilized for cross-modality alignment, as elaborated next. 

\subsubsection{Global Alignment}
\label{sec:align_GA}
To globally align image and textual representations, we first project local patch embeddings $\mathbf{f}^P_{i}$ to $\mathbf{\hat{f}}^P_{i}$ and average-pool them to generate projected global image embedding $\mathbf{\hat{f}}^G_{i}$. We then project the first \texttt{[CLS]} token, $\mathbf{t}^{C}_{i}$, of the sequence token representations $\mathbf{t}_{i}$ into the shared image-text embedding space and generate $\mathbf{\hat{t}}^{C}_{i}$ as global textual representation. 
We then follow a contrastive-learning strategy \cite{GLoRIA_ICCV21,MSCXR_ECCV22,granularity_NIPS22} to apply symmetric alignment loss to semantically align global image and text features:
\small\begin{multline}
\mathcal{L}_G = -\frac{1}{N} \sum^N_{i=1} (\log \frac{\exp{(\mathbf{\hat{f}}^G_i \cdot \mathbf{\hat{t}}^C_i} / \tau_G)}{\sum^N_{j=1}\exp{(\mathbf{\hat{f}}^G_i \cdot \mathbf{\hat{t}}^C_j} / \tau_G)} \\
+ \log \frac{\exp{(\mathbf{\hat{t}}^C_i} \cdot \mathbf{\hat{f}}^G_i / \tau_G)}{\sum^N_{j=1}\exp{(\mathbf{\hat{t}}^C_i} \cdot \mathbf{\hat{f}}^G_j / \tau_G)}),
\label{eq:GA}
\end{multline}
\normalsize
where $N$ is the number of image-text pairs in a training batch, $\tau_G$ is the scaling temperature parameter, and $\mathbf{\hat{f}}^G_i \cdot \mathbf{\hat{t}}^C_j$ is the cosine similarity between global image $\mathbf{\hat{f}}^G_i$ and textual embedding $\mathbf{\hat{t}}^C_j$. 
% The global alignment loss ensures semantic alignment of image and text representations in the joint embedding space.

\subsubsection{Local Alignment}
\label{sec:align_LA}
While global alignment with $\mathcal{L}_G$ semantically reconcile image and text representations, region-/word-level lesion-report association with finer granularity is crucial to localization-related downstream tasks like grounding and the proposed \problemname~\cite{MSCXR_ECCV22,GLoRIA_ICCV21,mutual_MICCAI21,jointlearn_ECCV22,granularity_NIPS22,muellerChEx_ECCV24,LIMITR_2023_ICCV}.
% pathologies typically only occupy small fractions of the CXR image and global alignment tends to overlook \textit{region-level} pathology-report correspondence. 
To this end, we propose a Context-Aware Alignment (CAA) module consisting of intra-modality and cross-modality correspondence learning in a cascaded way.% to learn region-/word-level aligned multimodal representations.

Specifically, given the sequence of token representations $\mathbf{t}_{i}$ generated by the text encoder from the input report $T_i$ (Section~\ref{sec:align_pre}), we first enforce self-attention \cite{bert_NAACL19} by computing the Multi-Head Attention (MHA) \cite{bert_NIPS17} where each token attends to every other token in the sequence, to capture intra-sequence relationships and highlight informative word tokens and generate initial self-attentive token features $\mathbf{t}^{\operatorname{SA}}_{i}$:
\small\begin{equation}
\mathbf{t}^{\operatorname{SA}}_{i}=\operatorname{Softmax}\Big(\sqrt{\frac{1}{d_k}} W_s^q\mathbf{t}_{i} \cdot W_s^k\mathbf{t}_{i}\Big) W_s^v\mathbf{t}_{i},
\label{eq:SA}
\end{equation}\normalsize
where $W_s^q$, $W_s^k$ and $W_s^v$ are learnable weight matrices specific to each attention head. We then perform a residual operation to compute final self-attentive token features as: $\mathbf{t}^{\operatorname{SA}'}_{i} = \mathbf{t}^{\operatorname{SA}}_{i} + \mathbf{t}_{i}$.

To align local representations across image and text modalities, we perform cross-attention with MHA by querying self-attentive token features $\mathbf{t}^{\operatorname{SA}'}_{i}$ with image patch features $\mathbf{f}^P_{i}$, to generate initial cross-attentive/modality token features $\mathbf{t}^{\operatorname{CA}}_{i}$ as:
\small\begin{equation}
\mathbf{t}^{\operatorname{CA}}_{i}=\operatorname{Softmax}\Big(\sqrt{\frac{1}{d_k}} W_c^q\mathbf{t}^{\operatorname{SA}'}_{i} \cdot W_c^k\mathbf{f}^P_{i}\Big) W_c^v\mathbf{f}^P_{i}.
\label{eq:CA}
\end{equation}\normalsize
Similarly, $W_c^q$, $W_c^k$ and $W_c^v$ are learnable weight matrices specific to each attention head. The residual operation is then performed to obtain final cross-modality token features, $\mathbf{t}^{\operatorname{CA}'}_{i} = \mathbf{t}^{\operatorname{CA}}_{i} + \mathbf{t}^{\operatorname{SA}}_{i}$, for subsequent computations. 

We subsequently project cross-modality token features and self-attentive token features to image-text joint space: $\mathbf{t}^{\operatorname{CA}'}_{i} \rightarrow \mathbf{\hat{t}}^{\operatorname{CA}'}_{i}$, $\mathbf{t}^{\operatorname{SA}'}_{i}\rightarrow \mathbf{\hat{t}}^{\operatorname{SA}'}_{i}$. Then we compute the local alignment loss to enforce \textit{weighted consistency} between intra-modality and inter-modality feature representations. Specifically, we adopt \texttt{[CLS]} token attention weights $\mathbf{w}_{\text{\tiny\texttt{[CLS]}}}$ (the attention scores when \texttt{[CLS]} token attends to other tokens in the sequence) to suppress less meaningful words in our application context (\eg, ``a", ``is") and highlight informative local descriptions (\eg, ``pleural effusion"), for targeted context-aware local alignment:
\small\begin{multline}
\mathcal{L}_l = -\frac{1}{N} \sum^N_{i=1} \{\log \frac{\exp{[\mathbf{w}_{\text{\tiny\texttt{[CLS]}}} \circ (\mathbf{\hat{t}}^{\operatorname{SA}'}_{i} \cdot \mathbf{\hat{t}}^{\operatorname{CA}'}_{i}} / \tau_L)]}{\sum^N_{j=1}\exp{[\mathbf{w}_{\text{\tiny\texttt{[CLS]}}} \circ (\mathbf{\hat{t}}^{\operatorname{SA}'}_{i} \cdot \mathbf{\hat{t}}^{\operatorname{CA}'}_{j}} / \tau_L)]} \\
+ \log \frac{\exp{[\mathbf{w}_{\text{\tiny\texttt{[CLS]}}} \circ (\mathbf{\hat{t}}^{\operatorname{CA}}_i} \cdot \mathbf{\hat{t}}^{\operatorname{SA}'}_{i} / \tau_L)]}{\sum^N_{j=1}\exp{[\mathbf{w}_{\text{\tiny\texttt{[CLS]}}} \circ (\mathbf{\hat{t}}^{\operatorname{CA}'}_i} \cdot \mathbf{\hat{t}}^{\operatorname{SA}'}_{j} / \tau_L)]}\}.
\label{eq:LA}
\end{multline}\normalsize
Similar to Equation~\ref{eq:GA}, $N$ is the number of image-report pair in a training batch, $\tau_L$ is the scaling temperature parameter. ``$\cdot$"  is the cosine similarity operator, and ``$\circ$" denotes Hadamard product. 
The Context-Aware Alignment (CAA) module is repeated $M$ times to compute the final multi-modal feature representations ($M=3$ in our implementation, \cf supplementary material for detailed discussions).

\begin{figure*}[h!]
    \centering
    \includegraphics[width=.95\linewidth]{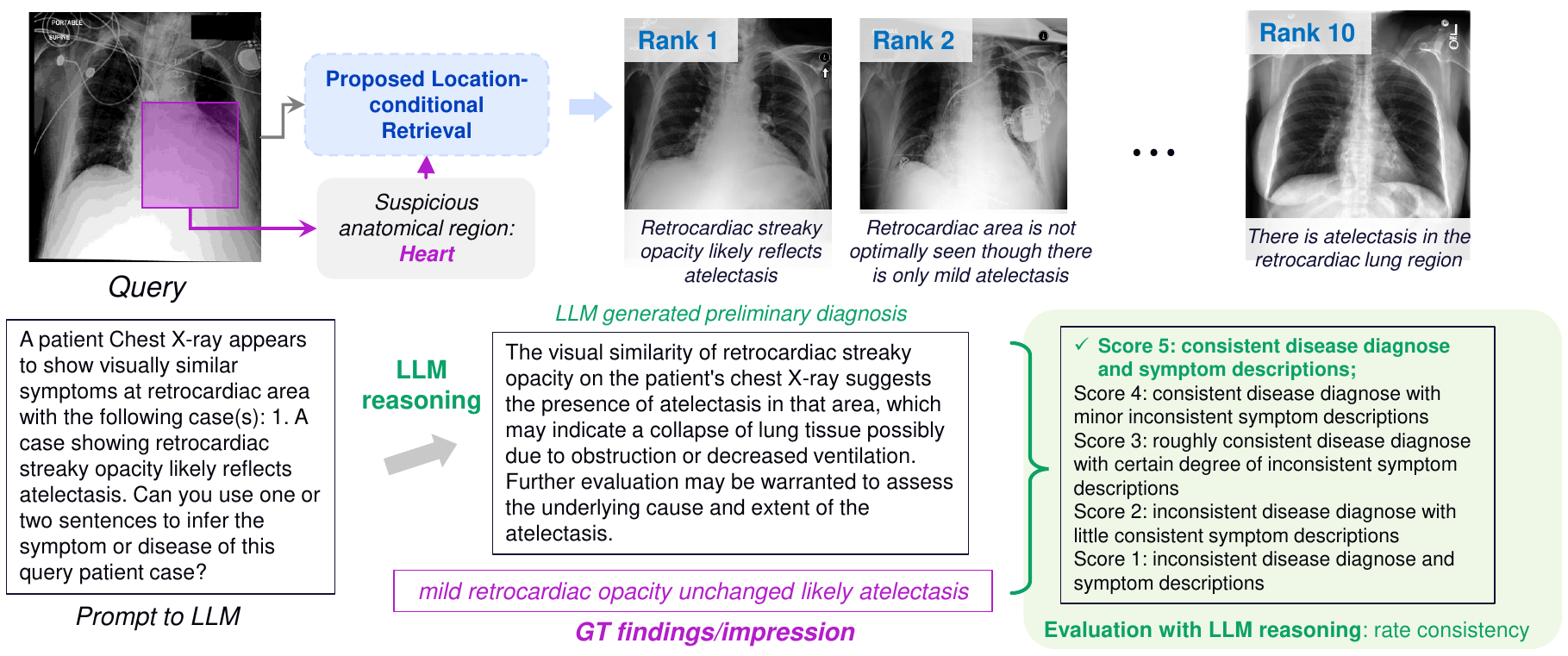}
    \caption{Proposed explanation generation pipeline based on location-conditioned multi-modal retrieval. Upper part shows retrieved patient gallery given the query image conditioned on ``Heart" region. Bottom part illustrates the explanation generation and evaluation pipeline.
    }
    \vspace{-0.4cm}
    \label{fig:explanation_pipeline}
\end{figure*}

\subsection{Location-conditioned Contrastive Learning}
\label{sec:method_triplet}
While global and local alignment mechanisms introduced in Section~\ref{sec:method_alignment} align visual and textual representations across different granularity levels, 
% in the unsupervised way, ensuring the generalizability and representation capability of the learned multi-modal embeddings. However, 
intra-pair contrastiveness is not fully explored during the unsupervised learning process. This may result in suboptimal performance in downstream tasks like classification or retrieval. 
One may perform conventional contrastive learning given disease labels (typically accessible for CXR datasets) to overcome this limitation, \ie, constructing positive and negative pairs with patient data labeled with same or different diseases. However, disease-level contrastive learning based on global image/text features bring further challenges, considering that: 1) medical data is typically multi-label data, \ie, a patient may commonly have concurrent diseases happening at different anatomical regions, so constructing positive/negative data pairs solely based on disease categories can easily confuse the model and compromise its performance; 2) lesions commonly span over tiny regions of a medical image, so contrastive learning based on global representations without explicit region-/location-conditioning is not optimal for accurate retrieval and query case understanding. 
% \mznote{rewritten, pls check}
% \mznote{However}, \jznote{I slightly revised this sentence} the local and pathology-level differences across various patient cases, which could significantly enhance location-aware image-text retrieval, are not fully leveraged. For example, a patient CXR image may indicate cardiomegaly at heart, accompanied with pleural effusions at lungs while other patient cases might not show cardiomegaly but only pleural effusions. Retrieval solely based on global image-text information can easily confuse these two cases with subtle visual and textual differences, thus result in suboptimal retrieval accuracy and affect subsequent diagnosis. Our objective is location-conditioned retrieval, where a radiologist can retrieve similar patient cases conditioned on anatomical regions. \Eg, for the aforementioned query patient case, if conditioned on heart, our model retrieves similar cases with cardiomegaly, while conditioned on lungs, we retrieve similar cases showing same pleural effusions symptom at lungs. In this case, our model can assist radiologists' diagnosis at pathology-region level, improving precision and interpretability. \mznote{I'll rewrite this para.}

To this end, we propose to enforce local/regional contrastive learning based on region-level disease labels as weak supervision, for improved location-conditioned retrieval (\problemname) performance. 
However, existing CXR datasets either contain only image-level disease labels without explicit region annotations like bounding boxes or segmentation masks~\cite{MIMIC_SD16,chexpert_aaai19,openI_dataset}; or have too few of them~\cite{MSCXR_ECCV22}, which may easily cause model overfitting and are thus not optimal for our study. Therefore, we propose to use general LLMs, \ie, GPT-4o~\cite{gpt4} to preprocess MIMIC-III/CXR~\cite{MIMIC_SD16} data and extract region-level information from the radiology reports. 
Specifically, we use GPT-4o to extract anatomical region information and corresponding symptom/disease descriptions from the report. 
\Eg, a CXR image with report stating ``linear opacity at the right lung base is suggestive of subsegmental atelectasis". ``right lung base" and ``subsegmental atelectasis" are extracted as the region and disease descriptions respectively. 
% a query has ``left lung" and ``severe pulmonary edema" as the region and disease descriptions respectively, 
We then form triplets based on region and disease labels, with positive pairs constructed as samples showing same symptom/disease at same anatomical region and negative pairs otherwise (different symptoms/diseases at same region or different symptoms/diseases at different regions, \cf supplementary material Section 1 for more detailed discussions).
% same anatomical region with different symptoms/diseases or different anatomical regions with same or different diseases. 
% \bpnote{a bit unclear maybe . How is the initial query obtained? Just sampled from the dataset? And are we assuming that all the CXR images are aligned when picking different regions/bboxes?} \mznote{a figure illustration is added in suppl., maybe refer to that figure here?} 
To enforce regional contrastive learning, we input the CXR images and region descriptions into the image and text encoders to generate region-conditioned cross-attentive features and minimize the triplet loss on positive/negative pairs as:
% \begin{equation}
% \mathcal{L}_{tr}= \frac{1}{N} \sum^N_{i=1} \max(\operatorname{d}(\mathbf{f}_i^{a},\mathbf{f}_i^{p})-\operatorname{d}(\mathbf{f}_i^{a},\mathbf{f}_i^{n}) + \alpha, 0).
% \label{eq:triplet}
% \end{equation}
% Here $N$ is the total number of triplet pairs in a sampled batch, $\operatorname{d}(\mathbf{f}_i^{a},\mathbf{f}_i^{p})$ is the $\ell_2$ distance between anchor and positive data pairs, $\operatorname{d}(\mathbf{f}_i^{a},\mathbf{f}_i^{n})$ is  the $\ell_2$ distance between anchor and negative data pairs 
% \bpnote{Could we just directly write ``$\ell_2(...)$ in the Equation rather than ``$d(...)$, to save space on explaining $d$ here?}, 
% $\alpha$ is the predefined margin hyperparameter. $\mathbf{f}_i$ is the cross-attentive feature vector from context-aware alignment (CAA) module for data $i$. \mznote{is it clear? should I use $\mathbf{f}$ or $\mathbf{\hat{t}}^{\operatorname{SA}'}_i$ for easier understanding?}
\begin{equation}
\mathcal{L}_{tr}= \frac{1}{N} \sum^N_{i=1} \max(\ell_2(\mathbf{\hat{t}}_i^{a},\mathbf{\hat{t}}_i^{p})-\ell_2(\mathbf{\hat{t}}_i^{a},\mathbf{\hat{t}}_i^{n}) + \alpha, 0).
\label{eq:triplet}
\end{equation}
Here $N$ is the total number of triplet pairs in a sampled batch, $\alpha$ is the predefined margin hyperparameter, and $\mathbf{\hat{t}}_i^a, \mathbf{\hat{t}}_i^p, \mathbf{\hat{t}}_i^n$ correspond to the cross-attentive feature vectors (\ie, $\mathbf{\hat{t}}^{\operatorname{SA}'}_i$) from the context-aware alignment module, respectively for the anchor, positive, and negative samples for triplet $i$.

\subsection{Overall Training and Interpretability}
\label{sec:method_exp}
We perform a two-stage training to learn our proposed RRA-VL. \Ie, we first finetune \modelname~with global alignment loss and local alignment loss (Section~\ref{sec:method_alignment}) as $\mathcal{L}_{a} = \mathcal{L}_G + \beta \mathcal{L}_l$, on MIMIC-III/CXR~\cite{MIMIC_SD16} dataset. 
% Specifically, we follow~\cite{MSCXR_ECCV22} and use IMPRESSION and FINDINGS sections in the report along with the CXR image to learn global-local aligned multi-modal representation.
We then train \modelname~with $\mathcal{L}_{tr}$ on processed MIMIC-III/CXR dataset (as illustrated in Section~\ref{sec:method_triplet}) with location-conditioned triplet pairs as the input to \modelname~to enforce regional contrastive learning. Dense annotations, \eg, bounding boxes or segmentation masks, are not required for training our \modelname. Refer to Section~\ref{sec:implementation} and supplementary material for additional implementation details.

Given a query CXR image and the suspicious anatomical region(s), our proposed \ours~pipeline can effectively retrieve similar cases from database conditioned on the querying region, interpretations and explanations can thus be automatically generated by referring to the retrieved patient gallery, as shown in Figure~\ref{fig:explanation_pipeline}. 
% similar or dissimilar patient cases across the database, with images and corresponding reports. 
Specifically, given the retrieved gallery with patients' metadata (including CXR image-report pairs) ranked with region-level similarities, we leverage LLMs (GPT-4o-mini~\cite{gpt4} in this example) to generate explanations given retrieved top-k similar case(s) based on their regional disease descriptions. \Eg, our proposed \ours~system retrieves a patient gallery given the query image in Figure~\ref{fig:explanation_pipeline}, when conditioned on suspicious anatomical region ``heart". We then prompt LLMs with the top-$k$ ($k$=1 in Figure~\ref{fig:explanation_pipeline} for illustration purpose) patient report(s) to generate explanations and preliminary diagnosis. Ground-truth reports of the query image can then be utilized for evaluations, which will be illustrated in Section~\ref{sec:eval_exp}.

\section{Experiments}
To evaluate the efficacy of our proposed \ours~system, we first validate the representation capability of the learned multi-modal visual-textual embedding by demonstrating state-of-the-art localization performance of our \modelname~model on phase grounding tasks. 
We then demonstrate the capability of \ours~in performing Location-Conditioned Multi-Modal Retrieval (\problemname), as well as conventional cross-modality retrieval tasks, showing superior accuracy when compared to existing medical VLM solutions. 
We finally study and evaluate explainability of the proposed \ours~compared to conventional image-text retrieval system. 
Implementation details and additional experimental results can be found in the supplementary material.

\subsection{Experimental Protocol}
\label{sec:implementation}
% \noindent \textbf{Implementation Details}. We adopt ResNet-50~\cite{resnet_he2015} as our image encoder and BERT~\cite{bert_NAACL19,bert_NIPS17} as our text encoder, initialized with weights pretrained from BioViL~\cite{MSCXR_ECCV22} following conventional multi-modal learning schemes~\cite{GLoRIA_ICCV21}. We perform a two-stage training scheme as introduced in Section~\ref{sec:method_exp}. Specifically, for first-stage learning, we input ``IMPRESSION" and ``FINDINGS" sections of a CXR report to the text encoder along with the CXR image (posteroanterior, PA view only) to the image encoder following similar strategy in~\cite{latent_NIPS21,MSCXR_ECCV22}, as ``IMPRESSION" and ``FINDINGS" sections of a CXR report typically contain detailed information of radiological interpretations from the radiologist. During second-stage training of \modelname, triplet pairs are generated online (\cf~Section~\ref{sec:method_triplet}) given location description and disease category labels. We adopt AdamW optimizer \bpnote{AdamW? add ref?} with learning rate $5e^{-6}$, weight decay $0.01$ for training our context-aware alignment (CAA) module. We use Adam optimizer with learning rate $5e^{-6}$, weight decay $1e^{-5}$ for finetuning the image and text encoder. Total training epoch is set to 10. We resize and center crop images to 512$\times$512 and perform random horizontal flips as image augmentations. \mznote{this para can be moved to suppl?}

\noindent\textbf{Datasets}.
a) \textbf{MIMIC-III/CXR}. We perform first-stage training on MIMIC-III/CXR~\cite{MIMIC_SD16} training set. 
b) \textbf{MIMIC-\textit{loc}}. For second-stage training of \modelname, 
% in order to generate location-conditioned image-text pair, we first use GPT-4~\cite{GPT4_openAI} to 
we preprocess original MIMIC-III/CXR dataset to extract region-level disease descriptions as illustrated in Section~\ref{sec:method_exp}, 
% \Eg, for a radiological report, the extracted anatomical location description can be ``Bilateral (both sides)" with disease category label ``Pleural Effusion". We then
and generate 15.9k/2.1k training/testing data samples with meaningful regional disease descriptions from MIMIC-III/CXR training/testing samples.
% \jznote{do we need to mention how (using what prompt) we generate the text using GPT? }. 
We refer to our re-curated MIMIC-III/CXR dataset as MIMIC-\textit{loc} in the following context for easier differentiation. MIMIC-\textit{loc} will be made publicly available upon acceptance of this paper.
c) \textbf{MS-CXR}. For evaluation of the finegrained localization performance of proposed \modelname~model, we perform phase grounding tasks on MS-CXR~\cite{MSCXR_ECCV22}. MS-CXR has local image region (bounding box) annotations with descriptions of eight radiology findings of 1,153 data samples, curated from MIMIC-III/CXR dataset. 
d) \textbf{CheXpert 5x200.} To evaluate the multi-modal representation capability of learned \modelname~on conventional class-based image-report retrieval task, \ie, given an query image, the accuracy of retrieved reports belong to the same category (disease) as the query image; and given a query report, the accuracy of retrieved images belong to the same category as the query report for report-image retrieval.
% \bpnote{Do we mean ``We evaluate the multi-modal representation capability of learned \modelname~on conventional image-report retrieval task. \Ie, we measure the accuracy of retrieved reports given an query image, and the accuracy of retrieved images given a query report (retrieval is deemed correct when text and image refer to the same disease)."?} \mznote{correct!}. 
We perform cross-domain retrieval evaluation on CheXpert 5x200 dataset. 
% , following the protocol introduced in~\cite{Zhang2020ContrastiveLO_ICLR20,GLoRIA_ICCV21}. 
CheXpert 5x200 dataset is recurated from CheXpert~\cite{chexpert_aaai19}, which contains 200 exclusively positive images for each of category: atelectasis, cardiomegaly, consolidation, edema and pleural effusion, to avoid multi-label confusions for retrieval tasks. Please note that we do not train on CheXpert/CheXpert 5x200 dataset but only perform cross-domain evaluations to validate the generalizability of the learned multi-modal representations of \modelname.

\noindent\textbf{Metrics} For phase grounding, we follow~\cite{MSCXR_ECCV22,LIMITR_2023_ICCV} and use mean Intersection over Union (mIoU) and Contrast-to-noise ratio (CNR) for evaluating localization accuracy. We use Rank@K~\cite{GLoRIA_ICCV21,LIMITR_2023_ICCV,granularity_NIPS22} to measure the percentage of top-$k$ retrieved items are correct matches for retrieval tasks and mean Average Precision (mAP) to evaluate ranking quality across all retrieved results. 

\subsection{Phase Grounding Evaluation}
\begin{figure*}[h!]
    \centering
    \includegraphics[width=\linewidth]{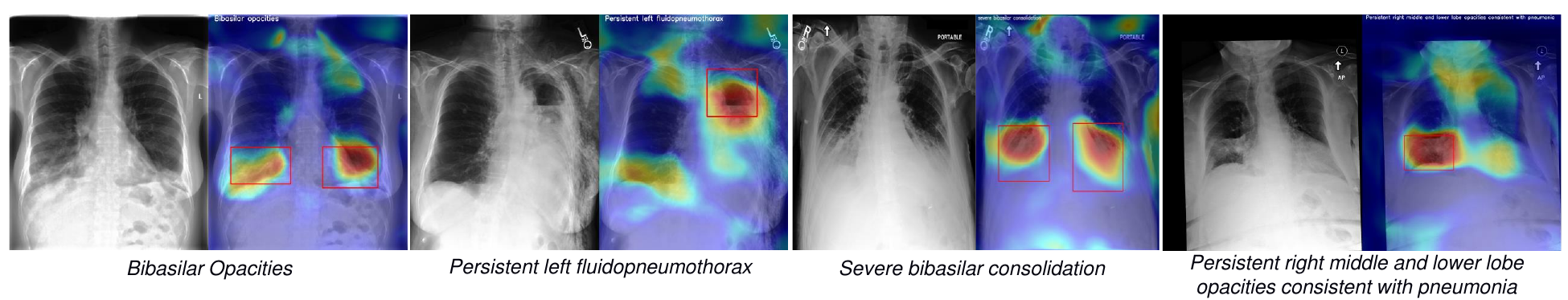}
    \vspace{-0.3cm}
    \caption{Visualization of phase grounding heatmaps of proposed \modelname~on MS-CXR~\cite{MSCXR_ECCV22}. Red boxes are ground-truth bounding boxes.}
    \vspace{-0.2cm}
    \label{fig:grounding_vis}
\end{figure*}

\noindent\textbf{Comparison to State-of-the-art.} 
To evaluate the localization capability of the proposed \modelname~model, we perform phase grounding evaluation on MS-CXR~\cite{MSCXR_ECCV22}, strictly following protocol proposed in \cite{MSCXR_ECCV22}. Specifically for each testing image-report pair in MS-CXR dataset, we input CXR image and corresponding phase descriptions to \modelname~and compute cosine similarities between projected image patch embeddings $\mathbf{\hat{f}}^P$ and cross-modality text features $\mathbf{\hat{t}}^{\operatorname{CA}'}$ to generate the similarity maps. We then upsample these maps to the original CXR image size~\cite{fu2024featup} and compare with the ground-truth bounding-box annotations for mIoU and CNR calculations.
Results presented in Table~\ref{tab:cnr_per_cat} and \ref{tab:grounding} confirm that the proposed \modelname~achieves superior phase grounding accuracy in terms of both mIoU and CNR compared to existing methods, demonstrating its strong multimodal representation capability benefited from the proposed multi-granularity alignment learning scheme. Visualizations are presented in Figure~\ref{fig:grounding_vis}. 

\begin{table*}
    \centering
    \caption{Comparison of proposed \modelname~to prior arts, evaluated by per-category Contrast-to-noise ratio (CNR) on MS-CXR~\cite{MSCXR_ECCV22} dataset.}
    \resizebox{\textwidth}{!}{
    \begin{tabular}{c|ccccccccc}
    \toprule
    Methods  &  Atelectasis &  Cardiomegaly &  Consolidation &  Lung opacity &  Edema &  Pneumonia &  Pneumothorax &  Pl. effusion &  Avg. \\ 
    \midrule
    ConVIRT~\cite{Zhang2020ContrastiveLO_ICLR20} & 0.86$\pm$.04 & 0.64$\pm$.06 & 1.25$\pm$.06 & 0.78$\pm$.07 & 0.68$\pm$.07 & 1.03$\pm$.05 & 0.28$\pm$.08 & 1.02$\pm$.03 & 0.818$\pm$.01 \\
    GLoRIA~\cite{GLoRIA_ICCV21} & 0.98$\pm$.04 & 0.53$\pm$.31 & 1.38$\pm$.03 & 1.05$\pm$.04 & 0.66$\pm$.03 & 1.18$\pm$.04 & 0.47$\pm$.02 & 1.20$\pm$.04 & 0.930$\pm$.03 \\
    BioViL~\cite{MSCXR_ECCV22} & 1.02$\pm$.06 & 0.63$\pm$.08 & 1.42$\pm$.02 & 1.05$\pm$.06 & 0.93$\pm$.03 & 1.27$\pm$.04 & 0.48$\pm$.06 & 1.40$\pm$.06 & 1.027$\pm$.02 \\
    BioViL-L~\cite{MSCXR_ECCV22} & 1.17$\pm$.04 & 0.95$\pm$.21 & 1.45$\pm$.03 & 1.19$\pm$.05 & 0.96$\pm$.05 & 1.19$\pm$.01 & 0.74$\pm$.05 & \textbf{1.50}$\pm$.03 & 1.142$\pm$.04 \\
    LIMITR~\cite{LIMITR_2023_ICCV} & 1.16$\pm$n/a & \textbf{1.18}$\pm$n/a & 1.37$\pm$n/a & 1.37$\pm$n/a & 1.05$\pm$n/a & 1.27$\pm$n/a & 1.01$\pm$n/a & 1.24$\pm$n/a & 1.206$\pm$n/a \\
    \midrule
    \textbf{RRA-VL(ours)} & \textbf{1.48$\pm$.01} & {1.07$\pm$.01} & \textbf{1.58$\pm$.02} & \textbf{1.51$\pm$.04} & \textbf{1.15$\pm$.04} & \textbf{1.41$\pm$.01} & \textbf{0.95$\pm$.03} & {1.40$\pm$.04} & \textbf{1.333$\pm$.03}\\
    %\cmidrule(lr){5-7}
    \bottomrule
    \end{tabular}}
    \vspace{-0.4cm}
    \label{tab:cnr_per_cat}%
\end{table*}

\begin{table}[t]
    \centering
    \caption{Phase grouding performance comparison of \modelname~to state-of-the-art on MS-CXR~\cite{MSCXR_ECCV22} dataset. Results are averaged over five runs with different random seeds following~\cite{MSCXR_ECCV22}.}
    \resizebox{.48\textwidth}{!}{
    \begin{tabular}{c|c|c|c}
    \toprule
    {Methods} & Venues  & mIoU & CNR \\ 
    \midrule
    {Clinical-BERT~\cite{ClinicalBERT}} & ClinicalNLP 2019 & 0.182 & 0.791 \\
    {BioViL~\cite{MSCXR_ECCV22}} & ECCV 2022 & 0.194 & 0.796 \\
    {ConVIRT~\cite{Zhang2020ContrastiveLO_ICLR20}} & ML Healthcare 2022 & - & 0.818 \\
    {BioViL+MLM~\cite{MSCXR_ECCV22}} & ECCV 2022 & 0.209 & 0.860 \\
    {GLoRIA~\cite{GLoRIA_ICCV21}} & ICCV 2021 & - & 0.930 \\
{BioViL+dropout~\cite{MSCXR_ECCV22}} & ECCV 2022 & 0.217 & 0.945 \\
    {BioViL+RSM~\cite{MSCXR_ECCV22}} & ECCV 2022 & 0.220 & 1.012 \\
    {BioViL+CXR-BERT~\cite{MSCXR_ECCV22}} & ECCV 2022 & 0.220 & 1.031 \\
    \midrule
    \textbf{RRA-VL(ours)} & - & \textbf{0.348} & \textbf{1.276} \\
    %\cmidrule(lr){5-7}
    \bottomrule
    \end{tabular}}
    \label{tab:grounding}%
\end{table}

\noindent\textbf{Ablation Study.}
We evaluate the role of global alignment and the context-aware alignment module including the intra-modality and inter-modality correspondence learning. From the ablation study shared in Table~\ref{tab:abl_grounding}, we observe that context-aware local alignment brings larger benefits compared to global alignment to the phase grounding performance, with residual operations (``+Residual") and weighted consistency (``+$\mathbf{w}_{\text{\tiny[CLS]}}$") further improving localization accuracy. 

\begin{table}[t]
    \centering
    \caption{Ablation study of \modelname~on MS-CXR~\cite{MSCXR_ECCV22}, evaluated on phase grounding tasks.}
    \vspace{-0.3cm}
    \resizebox{.45\textwidth}{!}{
    \begin{tabular}{c|c|c|c|c|c}
    \toprule
    \multirow{ 3}{*}{\textit{\textbf{Global}}} & \multicolumn{3}{c|}{\textit{\textbf{Local}}} & \multirow{3}{*}{mIoU} & \multirow{3}{*}{CNR} \\
    \cline{2-4}
    & \multirow{2}{*}{\parbox{1.2cm}{\centering Intra-Modality}} & \multicolumn{2}{c|}{Cross-Modality} & & \\
    \cline{3-4}
    & & +Residual & +$\mathbf{w}_{\text{\tiny[CLS]}}$ & & \\
    \midrule
    \xmark & \xmark & \xmark & \xmark & 0.246 & 1.014 \\
    \cmark & \xmark & \xmark  & \xmark & 0.279 & 1.178 \\
    \xmark & \cmark & \cmark & \cmark & 0.305 & 1.255 \\
    \cmark & \cmark & \xmark & \xmark & 0.293 & 1.249 \\
    \cmark & \cmark & \cmark & \xmark & 0.323 & 1.272 \\
    \cmark & \cmark & \xmark & \cmark & 0.319 & 1.281\\
    \cmark & \cmark & \cmark & \cmark & \textbf{0.350} & \textbf{1.283} \\
    %\cmidrule(lr){5-7}
    \bottomrule
    \end{tabular}}
    \vspace{-0.4cm}
    \label{tab:abl_grounding}%
\end{table}

\subsection{Multimodal Retrieval Evaluations}
\noindent\textbf{Location-conditioned Multimodal Retrieval (\problemname).}
To validate the proposed \modelname~in performing \problemname, we input each CXR image and anatomical location descriptions in MIMIC-\textit{loc} testing set, to our proposed \modelname~to generate 
% image global embeddings $\mathbf{\hat{f}}^G$ and 
cross-modality text embeddings $\mathbf{\hat{t}}^{\operatorname{CA}'}$ as multimodal embeddings for the target retrieval task. 
Note that for each testing query image, we only perform retrieval at anatomical region(s) where disease(s) exist, \ie, where the ground-truth disease labels are positive, for evaluation purposes.
% \mznote{is this sentence/explanation necessary?}
% Specifically, more than one image-text pair can be generated as each CXR image may have multiple anatomical regions associated with different symptoms/diseases. 
We evaluated Rank@K and mAP of \modelname~for \problemname, under two settings in Table~\ref{tab:retrieval_loc}. 
``Region-level" retrieval refers to the scenarios when a retrieved CXR data is counted as a correct match for Rank@K evaluation if both anatomical location and disease category are the same as the query data. 
% \bpnote{I am not sure if ``location-conditioned" is the best term. My understanding is that in both cases, the query is conditioned on location ; but in setup 1, the correctness of the retrieved location matters whereas it doesn't in setup 2 (?). Would "region-level" vs. "image/global-level" retrieval work?}
``Image/global-level" refers to a more relaxed setting where the retrieved items are counted as correct as long as the disease category is consistent with the query, no matter the anatomical regions. To the best of our knowledge, this is the first proposed pipeline for \problemname, and as such, it lacks baselines for performance comparison. To ensure a fair evaluation, we conduct the same experiments using publicly accessible BioViL~\cite{MSCXR_ECCV22} checkpoints. We show that the proposed \modelname~achieves significantly better \problemname~performance compared to BioViL~\cite{MSCXR_ECCV22}. 
On the other hand, Table~\ref{tab:retrieval_loc} shows that proposed \modelname~achieves satisfying retrieval performance with alignment loss only (``w/o $\mathcal{L}_{tr}$"), with further performance improvement when location-conditioned triplet mining is added. This aligns with the expectation that the proposed alignment scheme enforces context-aware multimodal representation learning, which enables \problemname, with regional contrastive learning further boosts \problemname~performance.

\begin{table*}
    \centering
    \caption{Class-based location-conditioned multi-modal retrieval evaluation of \modelname~on MIMIC-\textit{loc} test set.}
    \vspace{-0.3cm}
    % \scriptsize
    \resizebox{.9\textwidth}{!}{
    \begin{tabular}{c|c|c|c|c|c|c|c|c|c}
    \toprule
      \multicolumn{2}{c|}{\multirow{ 2}{*}{\large Methods}} & \multicolumn{4}{c|}{\textit{\textbf{Region-level}}} & \multicolumn{4}{c}{\textit{\textbf{Image/Global-level}}} \\
    \cline{3-10}
   \multicolumn{2}{c|}{}& Rank@1 & Rank@5 & Rank@10 & mAP & Rank@1 & Rank@5 & Rank@10 & mAP \\ 
    \midrule
    \multicolumn{2}{c|}{BioViL~\cite{MSCXR_ECCV22}}  & 11.58 & 40.68 & 57.53 & 9.66 & 23.95 & 65.58 & 83.68 & 20.04\\ 
    \midrule
    \multirow{ 2}{*}{\textbf{\modelname~(ours)}} & w/o $\mathcal{L}_{tr}$ & 61.63 & 82.16 & 87.47 & 45.37 & 65.74 & 86.11 & 91.16 & 45.86 \\
    & w. $\mathcal{L}_{tr}$ & 65.11 & 84.37 & 89.00 & 51.92 & 67.95 & 86.74 & 91.79 & 53.43 \\
    \bottomrule
    \end{tabular}
    }
    \vspace{-0.4cm}
    \label{tab:retrieval_loc}%
\end{table*}

\noindent\textbf{Conventional Class-based Retrieval.}
To evaluate the generalizability of proposed \modelname~for conventional cross-modality retrieval, we report Rank@K Image2Report and Report2Image retrieval performance on CheXpert 5x200~\cite{chexpert_aaai19,Zhang2020ContrastiveLO_ICLR20,GLoRIA_ICCV21} dataset in cross-domain setting. Table~\ref{tab:retrieval_cls} shows the proposed \modelname~can achieves better Image2Report retrieval accuracy with large margin \cf Rank@5 and Rank@10 metric, with on-par Report2Image retrieval performance compared to BioViL~\cite{MSCXR_ECCV22}. Additional in-domain retrieval evaluation can be found in supplementary material.

\begin{table}
    \centering
    \caption{Class-based retrieval evaluation of \modelname~on CheXpert 5x200 dataset~\cite{Zhang2020ContrastiveLO_ICLR20,GLoRIA_ICCV21} (cross-domain evaluation).}
    \vspace{-0.3cm}
    \resizebox{0.48\textwidth}{!}{
    \begin{tabular}{c|c|c|c|c}
    \toprule
    \multicolumn{2}{c|}{Methods}  & Rank@1 & Rank@5 & Rank@10 \\ 
    \midrule
    \multirow{ 2}{*}{Image2Report} & BioViL~\cite{MSCXR_ECCV22}  & 19.94 & 51.30 & 59.52 \\ 
    & \textbf{\modelname~(ours)} & \textbf{20.44} & \textbf{61.32} & \textbf{98.90} \\
    \midrule
    \multirow{ 2}{*}{Report2Image} & BioViL~\cite{MSCXR_ECCV22}  & \textbf{24.55} & 69.14 & 86.67 \\
    & \textbf{\modelname~(ours)} & 20.54 & \textbf{72.65} & \textbf{91.68} \\
    
    \bottomrule
    \end{tabular}}
    \vspace{-0.5cm}
    \label{tab:retrieval_cls}%
\end{table}

\subsection{Explainability}
\label{sec:eval_exp}
\begin{figure}[t]
    \centering
    \includegraphics[width=.95\linewidth]{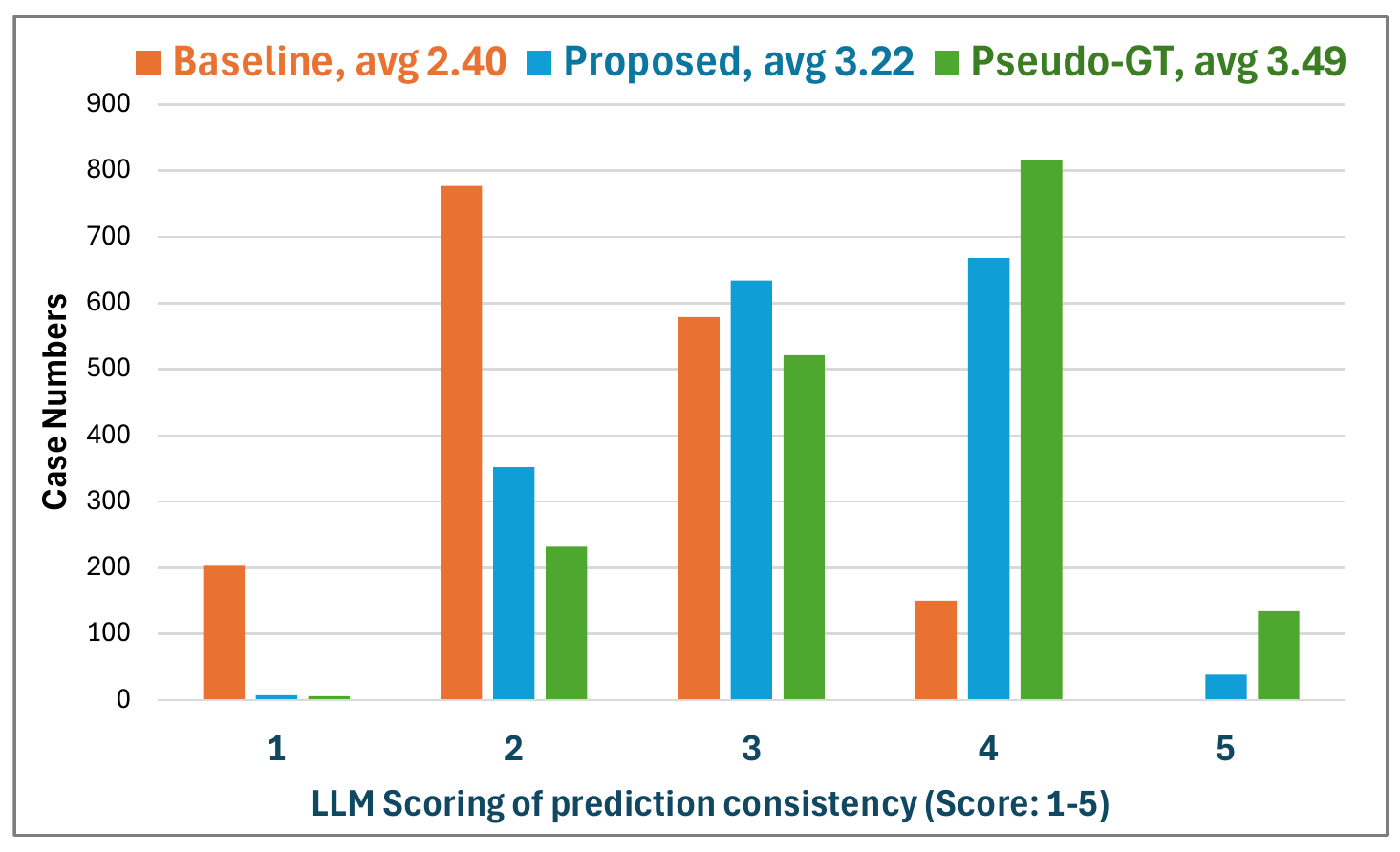}
    \caption{Statistics of rated consistency score (GPT-4o-mini) of generated explanations and GT descriptions, conditioned at different anatomical regions. Evaluation protocol is described in Figure~\ref{fig:explanation_pipeline}.
    }
    \vspace{-0.6cm}
    \label{fig:exp_rated_score_stats}
\end{figure}

As illustrated in Section~\ref{sec:method_exp}, our proposed \ours~pipeline naturally offers finer-level explainability, thanks to its capability in enabling retrieval conditioned at different anatomical regions. 
To evaluate the quality of generated preliminary diagnosis and explanations for each query case, we input the generated descriptions from LLMs and ground-truth region-level descriptions (available in MIMIC-loc dataset, \cf Section \ref{sec:implementation}), and prompt LLMs to rate a consistency score between them in the scale of 1-5 (Figure~\ref{fig:explanation_pipeline} bottom part and supplementary material gives a detailed explanation of the evaluation process and the metric definition of score 1-5). Figure~\ref{fig:exp_rated_score_stats} presents the average consistency score rated by GPT-4o-mini under 3 settings, ``Proposed" refers to the case where we use the top-1 retrieved patient report from proposed \ours~system to generate explanations and compared to ground-truth regional disease/symptom descriptions. ``Baseline" is similar to ``Proposed" but relies on global image representations instead of cross-attentive multi-modal features from \modelname~to perform retrieval and generate patient gallery (subsequent explanation generation and evaluation are the same as ``Proposed"). 
``Pseudo-GT" refers to the scenario where we use ground-truth top-1 match (as the retrieved top-1 match may not always be correct) to generate explanations and evaluate the consistency score accordingly. 
From Figure~\ref{fig:exp_rated_score_stats}, we can see that global-level retrieval (``Baseline") results in lower consistency scores with an average score of 2.40. This reflects the limitations of conventional global-level retrieval without explicit anatomical region conditioning, where inaccurate preliminary diagnosis may be generated due to suboptimal retrieval accuracy. 
On the other hand, our proposed \ours~pipeline (``Proposed") can generate reasonable explanations with average score 3.22, with small margin of degradation compared to ``Pseudo-GT", due to retrieval imperfection.
% \bpnote{Since all the methods (top1, top1\_baseline, top1\_CorrectMatch) use top-1 results (\ie, none uses top-$k$ with $k>1$), could we simplify the naming accordingly? I feel that just naming them "ours", "baseline", "pseudo-GT" could achieve the same meaning and shorten the legends / explanations ?}
% \mznote{@BP: good idea, updated}

\noindent \textbf{Discussion: Comparison to Image Report Generation.}
Conventional image report generation aims to generate a comprehensive report given a CXR image, and existing pipelines typically require finetuned/retrained text generators~\cite{chen_finegrained_ACL24,muellerChEx_ECCV24} with in-domain knowledge and full radiology reports as supervision. This can potentially lead to model overfitting with seen diseases and data modalities. Compared to image report generation pipelines, our proposed \ours~system can generate region-level disease or symptom descriptions without requiring any domain-specific text generator (only general LLMs like GPT-4o-mini), instead based on regional similarity/dissimilarity. As a result, our \ours~system is more flexible, and potentially more generalizable to unseen data and diseases. 
% \mznote{check?}

\section{Conclusion}
We propose an anatomical location-conditioned image-text retrieval (\ours) system, to retrieve relevant patient cases given query CXR image conditioned on anatomical region similarities. To perform \ours, we propose to learn a weakly-supervised \modelname~model with novel multi-granularity alignment and location-conditioned contrastive learning. Our proposed \ours~system is able to learn generalizable, well-aligned multimodal representations with superior localization capability, while enabling location-conditioned multimodal retrieval with fine explainability.

\clearpage
\setcounter{section}{0}
\twocolumn[{%
 \centering
 \LARGE Supplementary Material \\[1.5em]
 \normalsize
}]

\section{Implementation Details}
\noindent \textbf{Architecture}
We adopt ResNet-50~\cite{resnet_he2015} as our image encoder and BERT~\cite{bert_NAACL19,bert_NIPS17} as our text encoder, initialized with weights pretrained from BioViL~\cite{MSCXR_ECCV22} following conventional multi-modal learning schemes~\cite{GLoRIA_ICCV21}.

\begin{figure*}[h!]
\subfloat{\includegraphics[width=\linewidth]{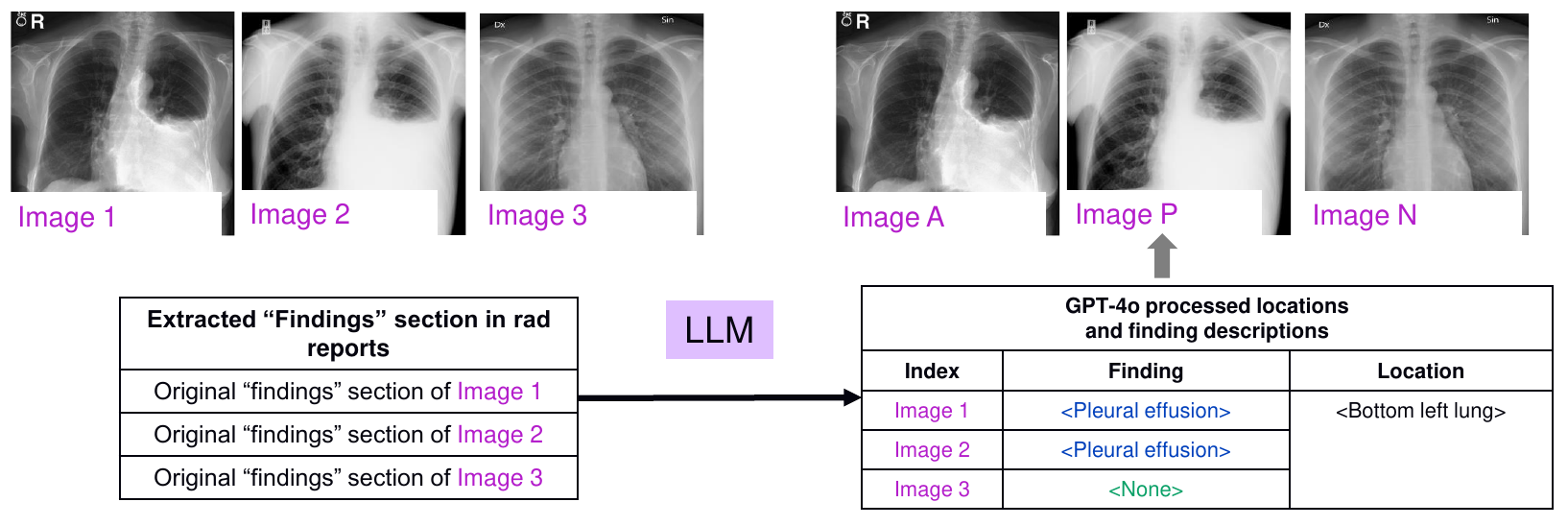}}\\
\caption{Illustration of the triplet sampling process for proposed location-conditioned contrastive learning. Image 1 and 2 are patient CXR images with ``pleural effusion" at anatomical region ``bottom left lung". Image 3 shows a patient CXR image with normal lungs.}
\label{fig:suppl_triplet}
\end{figure*}

\noindent \textbf{Training}
We perform a two-stage training scheme as introduced in Section 3.3 in the main paper. 
Specifically, for first-stage learning, we input ``IMPRESSION" and ``FINDINGS" sections of a CXR report to the text encoder along with the CXR image (posteroanterior, PA view only) to the image encoder following similar strategy in~\cite{latent_NIPS21,MSCXR_ECCV22}, as ``IMPRESSION" and ``FINDINGS" sections of a CXR report typically contain detailed information of radiological interpretations from the radiologist. 
During second-stage training of \modelname, triplet pairs are generated online (\cf~Section 3.3 in our main paper) given location description and disease category labels. 
Note that normal cases can be paired with normal/abnormal cases as positive/negative pair, regardless of the conditioning anatomical region. 
A figure illustration of the triplet sampling process can be found in Figure~\ref{fig:suppl_triplet}.

\noindent \textbf{Hyperpatameters}
We adopt AdamW optimizer~\cite{Loshchilov2017DecoupledWD} with learning rate $5e^{-6}$, weight decay $0.01$, $\beta=0.1$ (\cf Section 3.3 in the main paper) for training our context-aware alignment (CAA) module. We use Adam optimizer with learning rate $5e^{-6}$, weight decay $1e^{-5}$ for finetuning the image and text encoder. Total training epoch is set to 10. We resize and center crop images to 512$\times$512 and perform random horizontal flips as image augmentations. We empirically set $M=3$ (\cf Section 3.1 in the main paper) as we found that $M=3$ compromise the best between convergence and computation efficiency for CAA (model learning does not converge when $M<3$, while increasing $M$ does not improve performance on the other hand).

\noindent \textbf{Evaluation with LLM Reasoning}
For evaluation of consistency between generated explanations and ground-truth disease/symptom descriptions for each query (workflow illustrated in right bottom part of Figure 3 with experimental results shown in Figure 5), we prompt GPT-4o-mini with the following statement: ``Can you rate the consistency of the following two descriptions of a patient X-ray report in a scale of 1-5? Score 1 represents completely inconsistent disease diagnose and symptom descriptions. Score 2 represents inconsistent disease diagnose with little consistent symptom descriptions. Score 3 represents roughly consistent disease diagnose with certain degree of inconsistent symptom descriptions. Score 4 represents consistent disease diagnose with some inconsistent symptom descriptions. Score 5 represents completely consistent disease diagnose and symptom descriptions. Following are the two descriptions: A. (generated explanations). B. (ground-truth disease/symptom descriptions)". We then collect the returned score from GPT-4o-mini as the final rated consistency score for the query.

\section{In-Domain Multimodal Retrieval Results}
To demonstrate the capability of proposed \modelname~in performing conventional cross-modality retrieval. We follow protocol proposed in~\cite{GLoRIA_ICCV21,LIMITR_2023_ICCV} and perform in-domain cross-modality retrieval on MIMIC-III/CXR test set  and compare to existing medical VLM solutions. Specifically we perform compute cosine similarities between global image features $\mathbf{\hat{f}}^G$ and cross-attentive token features $\mathbf{t}^{\operatorname{CA}'}$ (\cf Figure 2 in the main paper) given CXR images and reports to rank similarities and obtain retrieval results. ``Image2Report" refers to cases where reports are being retrieved given query images, while ``Report2Image" denotes cases where images are being retrieved given query reports.
We report Rank@1, Rank@5 and Rank@10, and calculate the percentage of queries whose correct matches are successfully ranked within top-1, 5 and 10 in Table~\ref{tab:suppl_retrieval_cls}.
Table~\ref{tab:suppl_retrieval_cls} demonstrates that our proposed \modelname~achieves state-of-the-art Rank@5 and Rank@10 with noticable margin for both Image-Report retrieval and Report-Image retrieval.

\begin{table}[h!]
    \centering
    \caption{In-domain multimodal retrieval evaluation of \modelname~on MIMIC-III dataset~\cite{MIMIC_SD16}, compared to prior arts.}
    \resizebox{0.48\textwidth}{!}{
    \begin{tabular}{c|ccc|ccc}
    \toprule
    \multirow{ 2}{*}{Methods} & \multicolumn{3}{c|}{{\textbf{Image2Report}}} & \multicolumn{3}{c}{{\textbf{Report2Image}}} \\
    & R@1 & R@5 & R@10 & R@1 & R@5 & R@10 \\ 
    \midrule
    MOTOR~\cite{Lin2023TowardsMA} & 10.96 & 31.93 & 42.90 & 12.00 & 33.10 & 44.32 \\
    Im2Cap~\cite{Fang2014FromCT} & 18.60 & 43.10 & 56.10 & 18.13 & 43.20 & 55.97 \\
    JoImTeR-Net~\cite{JoImTeR-Net_MLMI21} & 18.93 & 46.20 & 58.67 & 19.07 & 45.27 & 58.50 \\
    MGCA~\cite{granularity_NIPS22} & 25.80 & 51.90 & 62.10 & 27.90 & 51.20 & 61.60\\
    ConVIRT~\cite{Zhang2020ContrastiveLO_ICLR20} & 30.10 & 53.90 & 63.80 & 29.20 & 54.70 & 64.40\\
    GLoRIA~\cite{GLoRIA_ICCV21} & 30.30 & 57.50 & 66.50 & 24.00 & 51.80 & 62.80\\
    LIMITR~\cite{LIMITR_2023_ICCV} & 36.10 & 59.10 & 69.10 & 36.40 & 60.70 & 70.50 \\
    LIMITR(\small{+LT\&PE})~\cite{LIMITR_2023_ICCV} & \textbf{39.70} & 63.20 & 71.70 & 37.70 & 62.10 & 71.30 \\
    \midrule
    \textbf{\modelname~(ours)} & 38.64 & \textbf{74.27} & \textbf{82.60} & \textbf{39.94} & \textbf{62.86} & \textbf{86.39} \\
    
    \bottomrule
    \end{tabular}}
    \label{tab:suppl_retrieval_cls}%
\end{table}

\section{Additional Visualizations}

\subsection{Phase Grounding}
\begin{figure*}[h!]
\subfloat{\includegraphics[width=\linewidth]{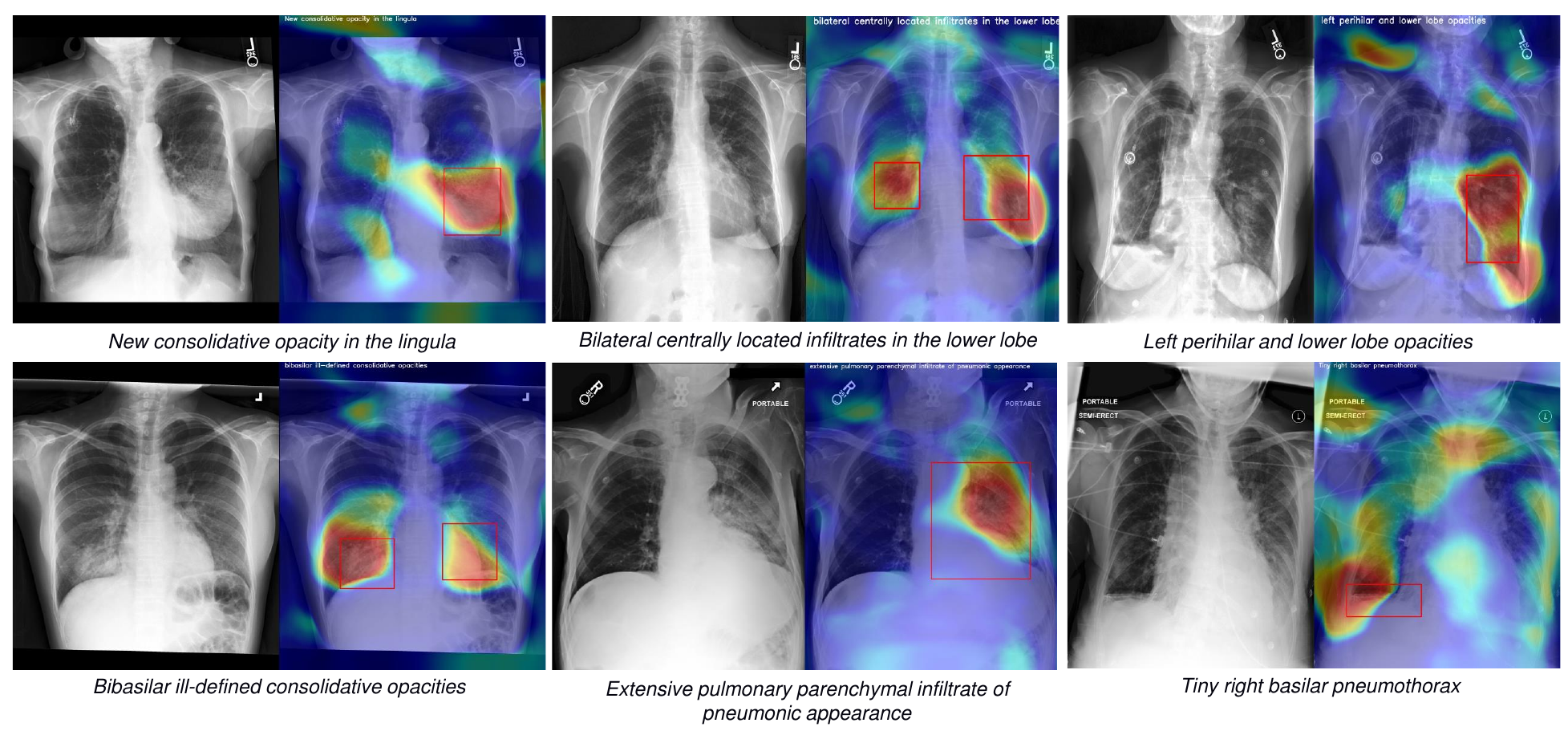}}\\
\vspace{-1em}
\subfloat{\includegraphics[width=\linewidth]{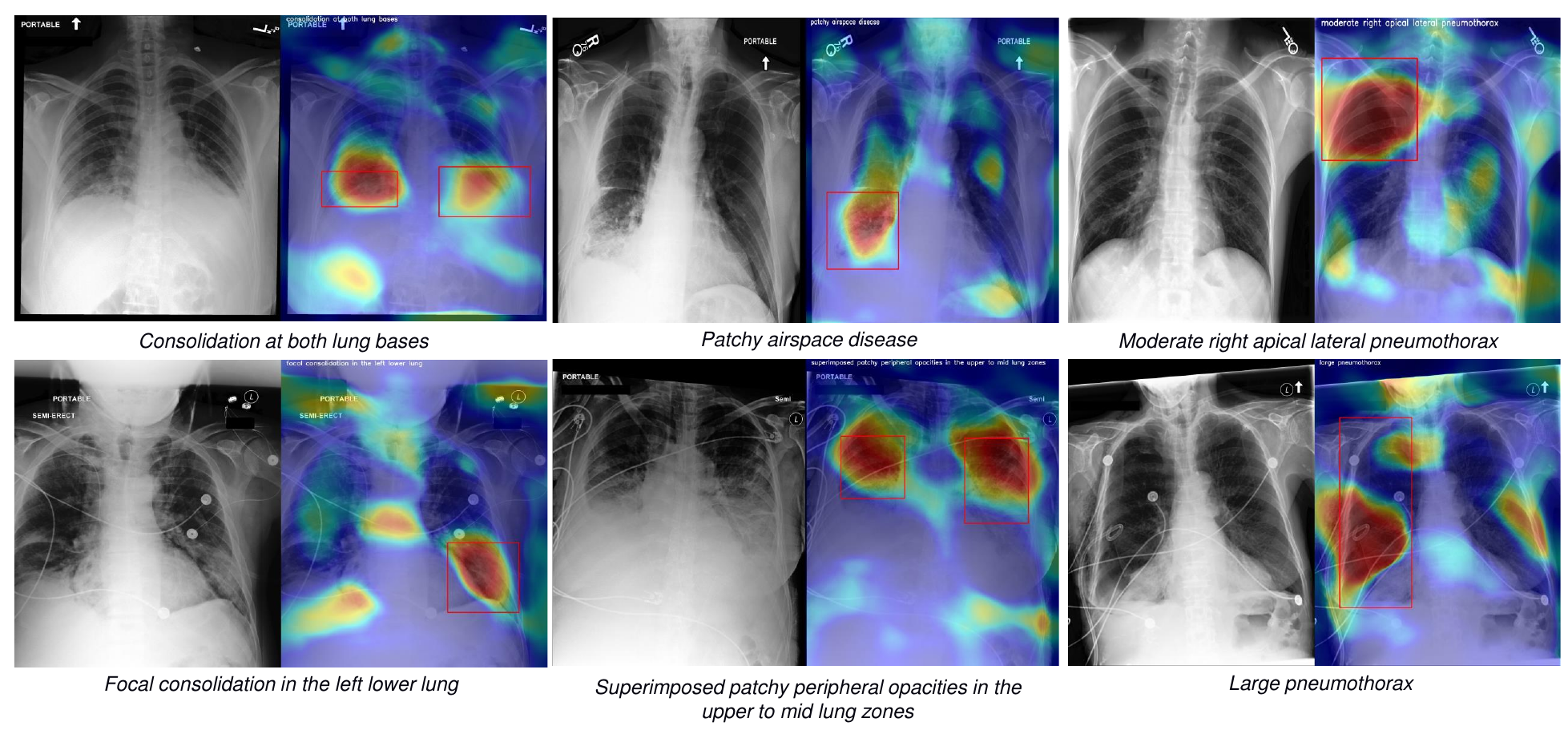}}\\
\vspace{-1em}
\subfloat{\includegraphics[width=\linewidth]{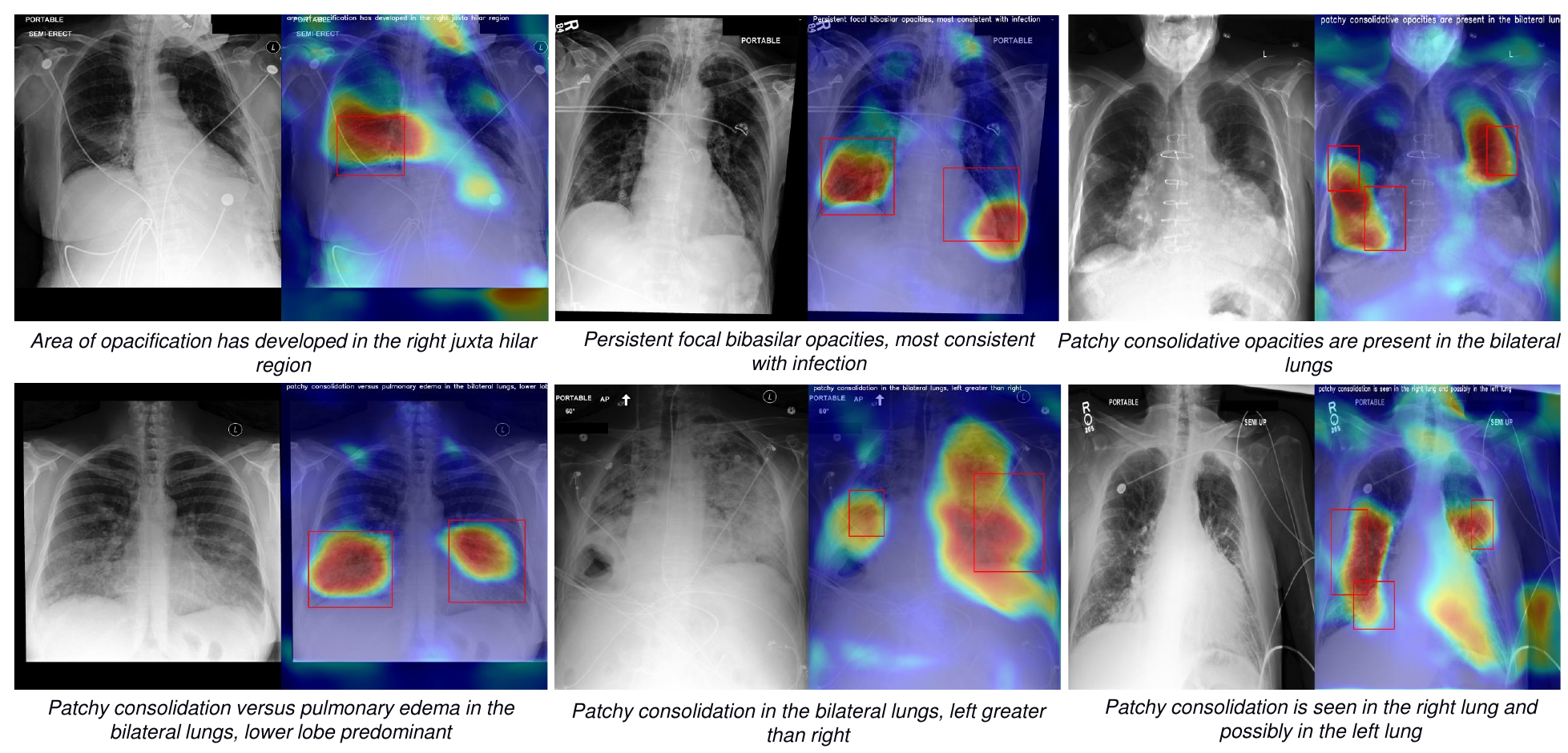}}
\caption{Visualization of phase grounding heatmaps of proposed \modelname~on MS-CXR~\cite{MSCXR_ECCV22}. Red boxes are ground-truth bounding boxes.}
\label{fig:suppl_grounding1}
\end{figure*}

\begin{figure*}[h!]
\subfloat{\includegraphics[width=\linewidth]{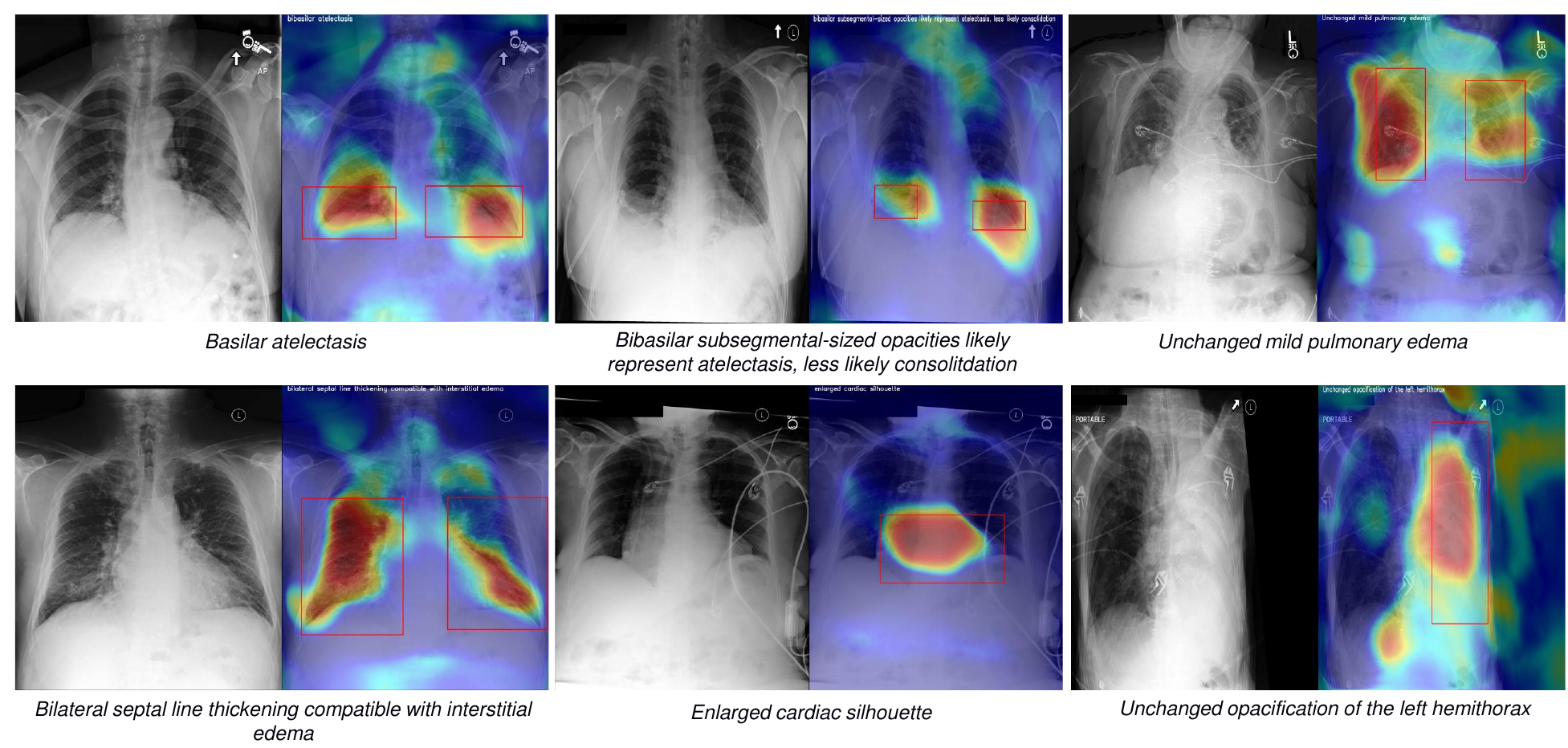}} 
\caption{Visualization of phase grounding heatmaps of proposed \modelname~on MS-CXR~\cite{MSCXR_ECCV22} (cont.). Red boxes are ground-truth bounding boxes.}
\label{fig:suppl_grounding2}
\end{figure*}

We show additional visualizations of phase grounding heatmaps (corresponding to Figure 4 in the main paper) of proposed \modelname~on MS-CXR~\cite{MSCXR_ECCV22} in Figure \ref{fig:suppl_grounding1} and Figure \ref{fig:suppl_grounding2}.

\subsection{Location-Conditioned Multimodal Retrieval and Explainability}
\begin{figure*}[h!]
\centering
\subfloat{\includegraphics[width=\linewidth]{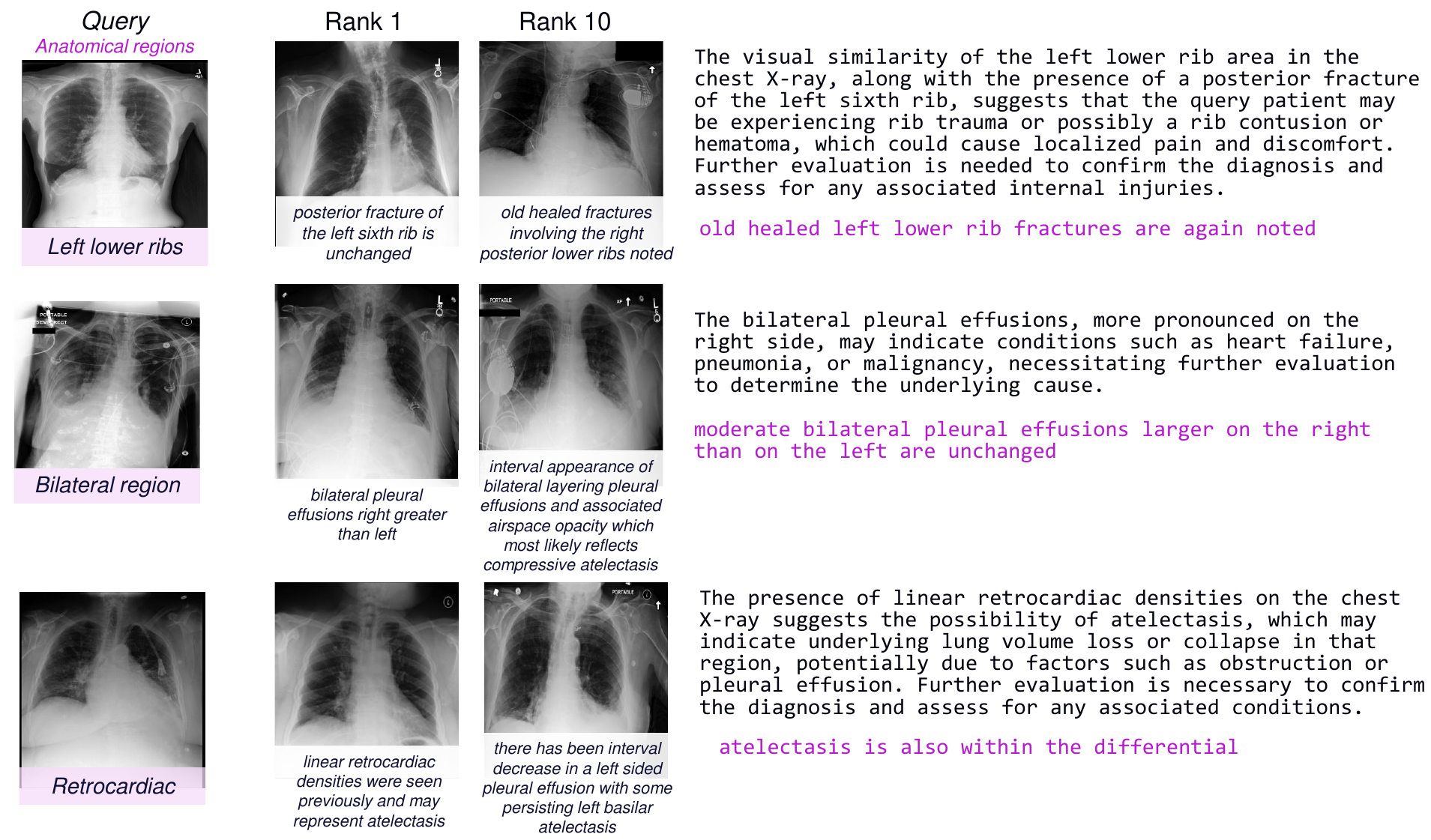}} \\
\vspace{0.2em}
\subfloat{\hspace{.2em}\includegraphics[width=\linewidth]{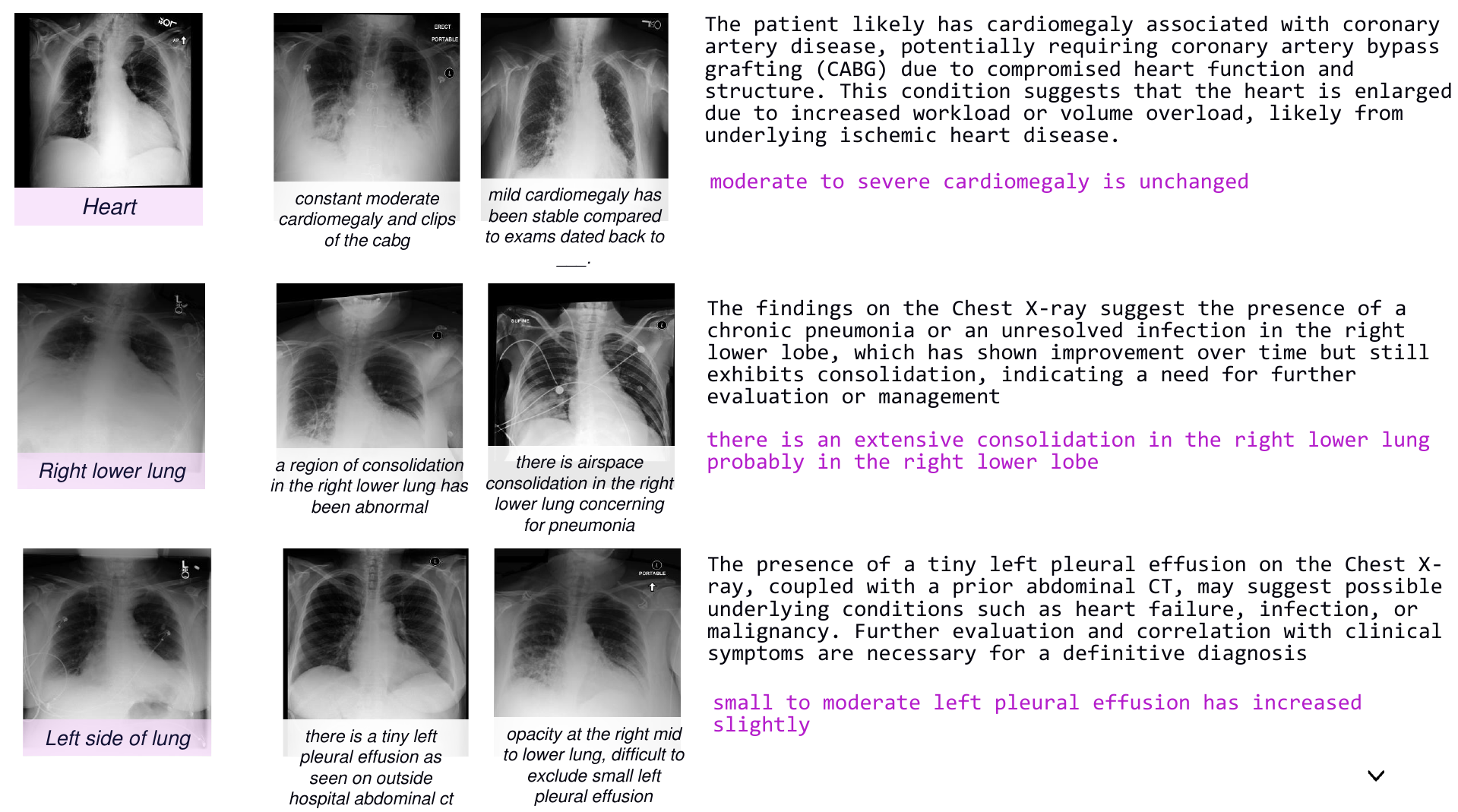}} 
\caption{Visualization of generated explanations and preliminary diagnosis from our proposed \ours~system. For each row, on the left we show one query CXR image with conditioned anatomical region (highlighted in purple boxes), retrieved top-1 and top-10 patient CXR image with report. On the right side, we show generated preliminary diagnosis (in black texts) and groundtruth disease/symptom descriptions (in purple texts) at the conditioned anatomical region accordingly.}
\label{fig:suppl_ICL1}
\end{figure*}

\begin{figure*}[h!]
\subfloat{\includegraphics[width=\linewidth]{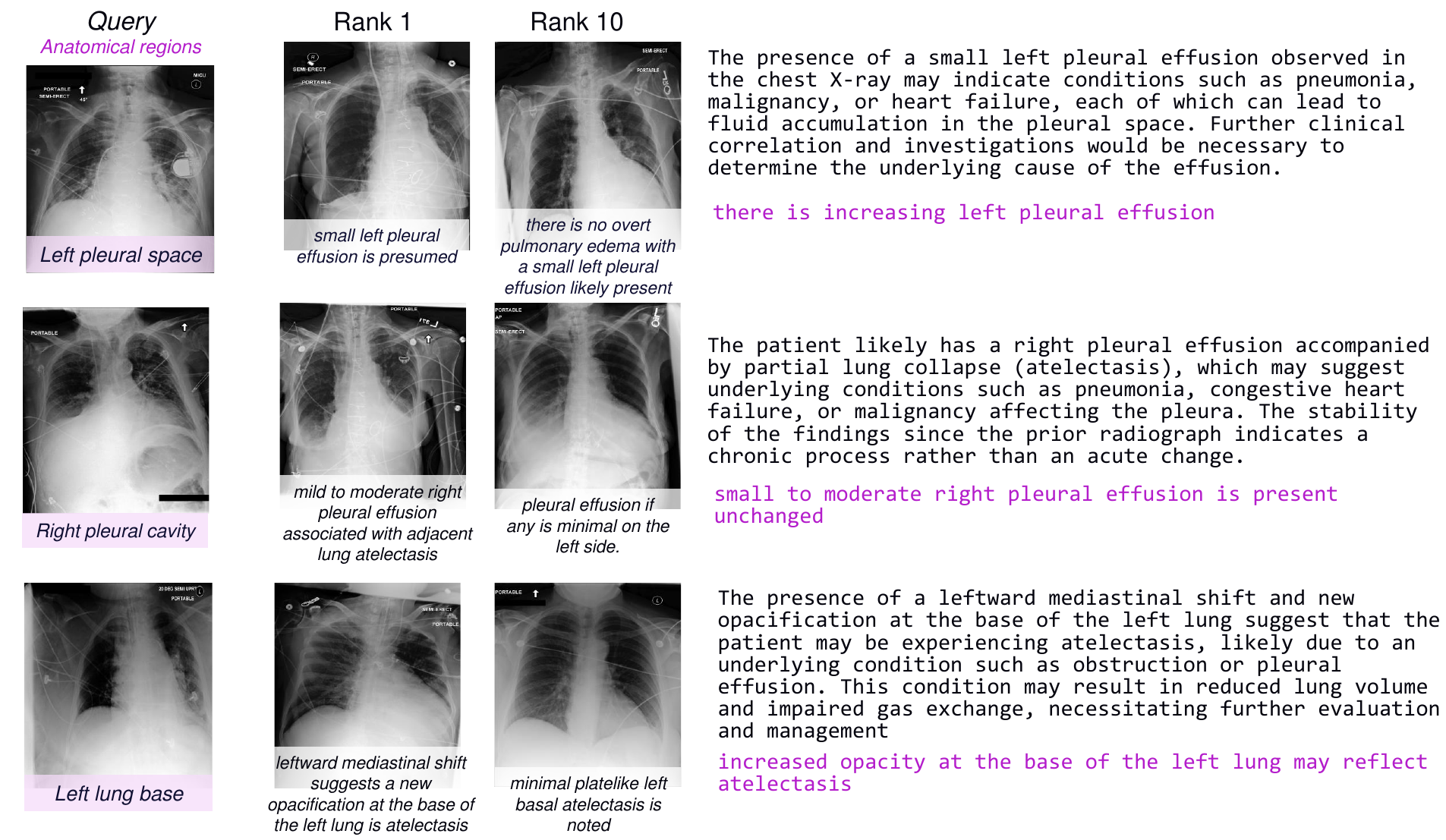}} \\
\subfloat{\hspace{.5em}\includegraphics[width=\linewidth]{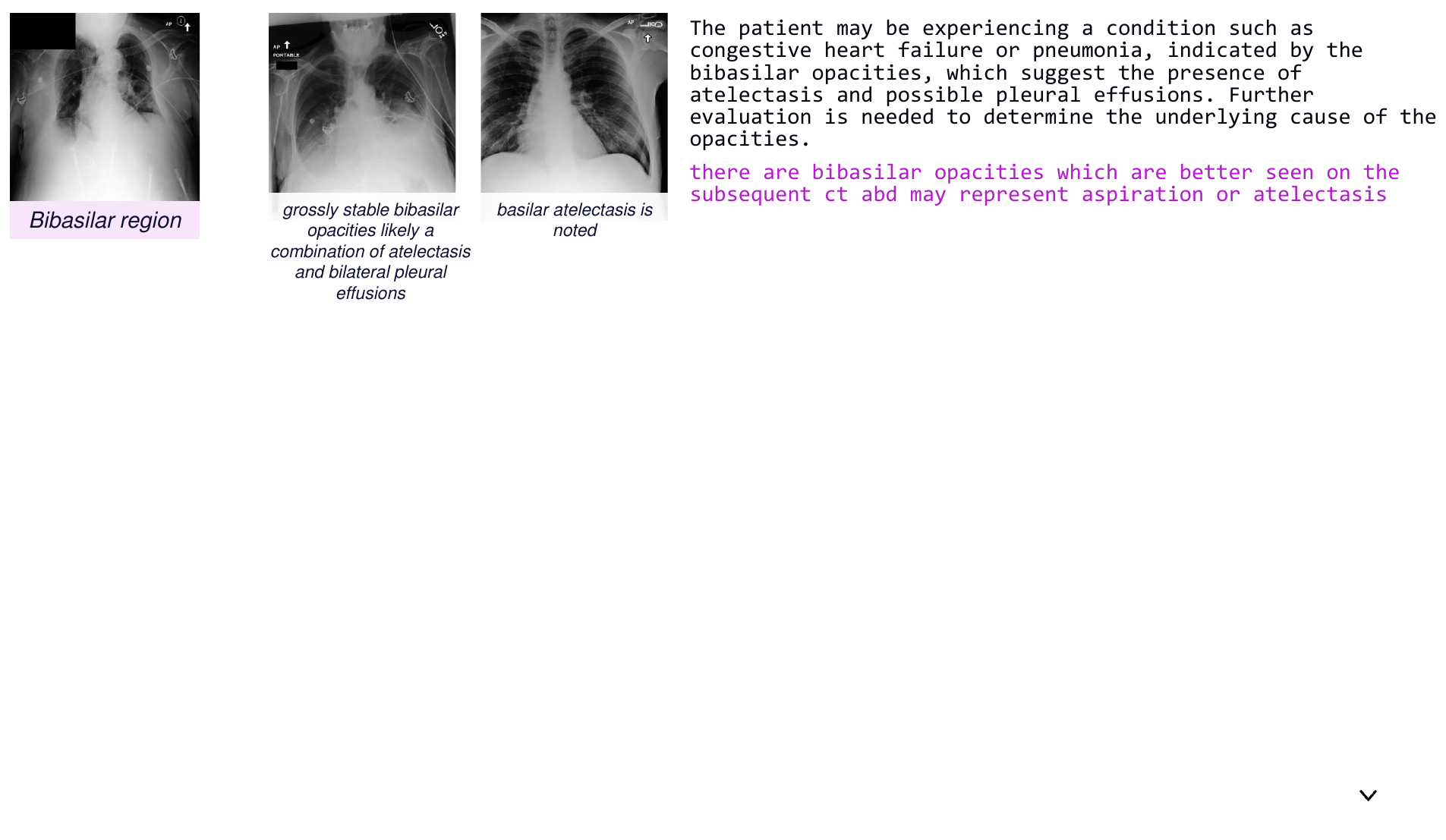}}\\
\caption{Visualization of generated explanations and preliminary diagnosis from our proposed \ours~system (cont.).}
\label{fig:suppl_ICL2}
\end{figure*}

We show additional visualization results of proposed location-conditioned multimodal retrieval on MIMIC-\textit{loc} (\cf Section 3.3 and 4.1 in the main paper) test set. We further demonstrate the explainability of the proposed \ours~system in providing fine-level explainability and preliminary diagnosis. Visualizations are shown in Figure \ref{fig:suppl_ICL1} and Figure \ref{fig:suppl_ICL2}.

\clearpage

{
    \small
    \bibliographystyle{ieeenat_fullname}
    \bibliography{references}
}

\end{document}